\documentclass[runningheads]{llncs}

\usepackage{eccv}

\usepackage{eccvabbrv}

\usepackage{graphicx}
\usepackage{booktabs}

\usepackage{multirow}
\usepackage{graphicx}
\usepackage{bbding}

\usepackage[accsupp]{axessibility}  

\usepackage{hyperref}

\usepackage{orcidlink}

\usepackage{wrapfig}

\begin{document}

\title{OMG: Occlusion-friendly Personalized Multi-concept Generation in Diffusion Models} 

\titlerunning{OMG: Occlusion-friendly Personalized Multi-concept Generation}


\author{
Zhe Kong\inst{1} \and 
Yong Zhang\inst{2}\Envelope \and
Tianyu Yang\inst{3} \and
Tao Wang\inst{4} \and
Kaihao Zhang\inst{5} \and
Bizhu Wu\inst{6} \and
Guanying Chen\inst{1} \and
Wei Liu\inst{2} \and
Wenhan Luo\inst{1,7}\Envelope
}

\authorrunning{Zhe. K et al.}

\institute{\textsuperscript{\rm 1}Shenzhen Campus of Sun Yat-sen University, \textsuperscript{\rm 2}Tencent AI Lab, \textsuperscript{\rm 3}International Digital Economy Academy, 
\textsuperscript{\rm 4}Nanjing University, \textsuperscript{\rm 5}Harbin Institute of Technology, Shenzhen, \textsuperscript{\rm 6}Shenzhen University,
\textsuperscript{\rm 7}The Hong Kong University of Science and Technology \\
Homepage: \url{https://kongzhecn.github.io/omg-project/} \\
Code: \url{https://github.com/kongzhecn/OMG/} }

\maketitle

\begin{figure}
\centering
\includegraphics[width=.85\linewidth]{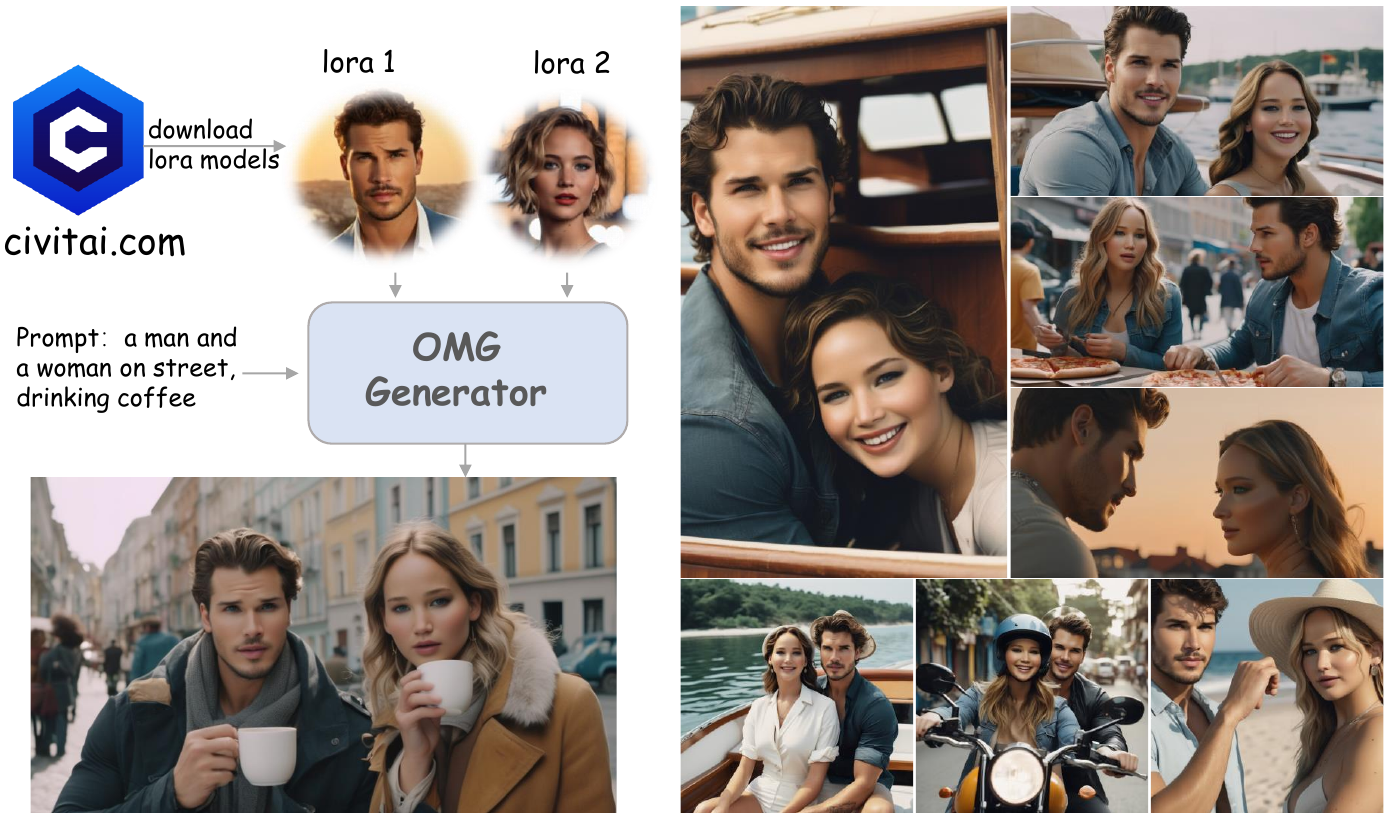}
\caption{We present OMG, an occlusion-friendly method for multi-concept personalization with strong identity preservation and harmonious illumination. The visual examples are generated by using LoRA models downloaded from \href{https://civitai.com/}{civitai.com}. 
}
\label{fig:example}
\end{figure}



\begin{abstract}
Personalization is an important topic in text-to-image generation, especially the challenging multi-concept personalization. Current multi-concept methods are struggling with identity preservation, occlusion, and the harmony between foreground and background. In this work, we propose OMG, an occlusion-friendly personalized generation framework designed to seamlessly integrate multiple concepts within a single image. We propose a novel two-stage sampling solution. The first stage takes charge of layout generation and visual comprehension information collection for handling occlusions. The second one utilizes the acquired visual comprehension information and the designed noise blending to integrate multiple concepts while considering occlusions. We also observe that the initiation denoising timestep for noise blending is the key to identity preservation and layout. Moreover, our method can be combined with various single-concept models, such as LoRA and InstantID without additional tuning. Especially, LoRA models on \href{https://civitai.com/}{civitai.com} can be exploited directly. Extensive experiments demonstrate that OMG exhibits superior performance in multi-concept personalization.

    
  \keywords{Image Generation \and Image Customization \and Diffusion Model}
\end{abstract}

\section{Introduction}
\label{sec:intro}

\begin{figure}[t]
    \centering
    \includegraphics[width=.85\textwidth]{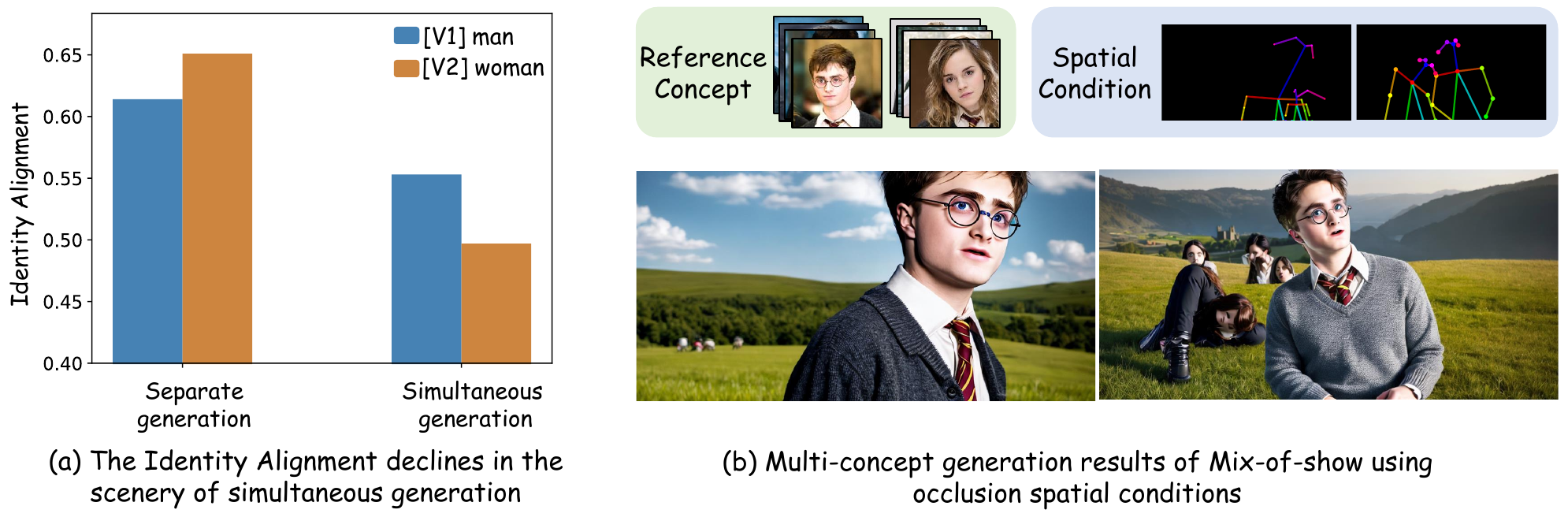}
    \caption{Existing methods face identity degradation and occlusion problems. (a) Given two text prompts with identifiers, ``A $[v1]$ man'' and ``A $[v2]$ woman'', we generate $100$ images for the two concepts separately (separate generation) and calculate the Identity Alignment between generated images and reference images. Subsequently, we employ another text prompt, ``A $[v1]$ man and a $[v2]$ woman'', to randomly generate $100$ images containing both concepts simultaneously (simultaneous generation) and calculate Identity Alignment. We find that the simultaneous generation of two concepts leads to the decline of Identity Alignment, resulting in identity degradation. (b) Given spatial conditions with occlusion between concepts, the Mix-of-show \cite{gu2023mix} cannot generate an integrity image and encounters an identity degradation problem.}
    \label{fig:motivation}
\end{figure}

Personalized text-to-image generation is a promising path to realize identity-consistent story visualization. 
Numerous methods have been proposed for single-concept personalization, such as DreamBooth \cite{ruiz2023dreambooth}, Textual Inversion \cite{gal2022image} and LoRA \cite{hu2021lora}, showcasing their efficacy in achieving high-quality results. While excelling in single-concept personalization, these methods encounter challenges related to identity degradation when tasked with generating a single image encompassing multiple concepts, as shown in Fig.~\ref{fig:motivation} (a).

Several multi-concept personalization methods have been proposed \cite{kumari2023multi, tunanyan2023multi,liu2023cones,gong2023talecrafter}, but they still encounter identity degradation problems when generating multiple concepts. 
Mix-of-show \cite{gu2023mix} can generate multi-concepts with realistic identity, but it cannot handle occlusion between concepts. Specifically, the method \cite{gu2023mix} adopts a regionally controllable sampling method, where each timestep injecting region prompts through regional-aware cross-attention. In cases where the concept regions experience occlusion, the final prediction results for these occluded regions are determined by a straightforward linear addition of the cross-attention results from multiple local sample regions. This simplistic approach leads to inaccurate predictions within the occluded regions, resulting in layout conflicts and identity degradation, as shown in Fig.~\ref{fig:motivation} (b). Besides, there is disharmony between the foreground and background, leading to unnatural illumination in the image.
Additionally, methods \cite{gu2023mix,kumari2023multi} aim at merging two concepts into one diffusion model, which is computationally inefficient.

To address the aforementioned issues, we propose OMG, an occlusion-friendly personalized image generation framework designed to seamlessly integrate multiple concepts within a single image. Unlike other customization methods, our two-stage approach employs latent-level and attention-level layout control to tackle occlusion issues during multiple concept customization. 
The first stage generates an image with coherent layouts based on user-provided text prompts, without considering personalization. During this stage, additional visual comprehension information such as attention maps and concept masks is acquired through the first stage of sampling. In the second stage, concepts are injected into specific regions by leveraging the preserved visual comprehension information. During sampling, as illustrated in Fig.~\ref{fig:motivation} (a), simultaneously generating two concepts in one image results in significant identity degradation. To address this limitation, we propose a concept noise blending strategy to merge multiple noises from different single-concept models during sampling. In each timestep, different single-concept models only control the generation of one specific region, effectively mitigating identity degradation problems during the multiple-concept sample process.
Additionally, we find that the disharmony problem can be solved by controlling the initiation timestep of concept noise blending. Differing from Custom Diffusion \cite{kumari2023multi} and Mix-of-show \cite{gu2023mix}, which require additional training or model optimization to merge multiple concepts into one model, the proposed OMG method can generate an image with multiple concepts directly by utilizing multiple single-concept models derived from the community (\textit{e.g.,} civitai.com) in a plug-and-play manner, without additional tuning. It is computationally efficient and significantly alleviates the time-consuming problem. Extensive experiments and comparison results with other methods demonstrate its superiority.
Our contributions are summarized as follows:
\begin{itemize}
    \item 
    We propose a novel two-stage framework for multi-concept customization. Our approach can generate an occlusion-friendly personalized image with strong identity preservation and harmonious illumination.
    \item We propose a Concept Noise Blending strategy to merge multiple noises from different single-concept models at both latent and attention levels. It mitigates identity degradation of the multi-concept generation and can be easily combined with different personalization frameworks such as LoRA or InstantID in a tuning-free plug-and-play manner.
    \item Extensive evaluations demonstrate the effectiveness of our proposed method.
\end{itemize}

\section{Related Work}
\label{sec:related}

\noindent\textbf{Text-to-Image (T2I) Synthesis.}
Text-to-image synthesis involves the task of generating realistic and diverse images from text prompts. Recently, diffusion models \cite{ho2020denoising, song2020denoising} have demonstrated remarkable progress, attributed to large-scale training datasets like Laion-400M \cite{schuhmann2021laion} and Conceptual-12M \cite{changpinyo2021conceptual}. Several text-to-image models, including SDXL \cite{zhou2023customization}, Imagen \cite{saharia2022photorealistic}, and DALL·E 3 \cite{betker2023improving}, have shown significant performance improvements. 

\noindent\textbf{Single-Concept Customization.}
Early image personalization approaches focus on expanding or fine-tuning the language vision dictionary of T2I diffusion models to associate new concepts with a limited set of subjects, achieved through the fine-tuning of pre-trained T2I models. Optimization-based methods, such as diffusion model based ones \cite{choi2023custom, hao2023vico, he2023data, ruiz2023dreambooth, ruiz2023hyperdreambooth, smith2023continual, chae2023instructbooth}, or special textual embeddings \cite{alaluf2023neural, gal2022image, vinker2023concept, voynov2023p+, pang2023cross, zhang2023compositional, zhao2023catversion}, learn new concepts to describe target concepts. To reduce the trainable parameters, recent advancements have seen the adoption of Low-Rank Adaptation (LoRA) methods \cite{hu2021lora, tewel2023key} in concept customization.
Moreover, studies \cite{arar2023domain, chen2023subject, gal2023encoder, ma2023unified, shi2023instantbooth, zhou2023enhancing, wei2023elite, gal2023designing, wang2024instantid, li2023photomaker, yan2023facestudio, chen2023anydoor, zhou2023customization, shi2023instantbooth} have recently explored training additional modules for mapping concepts to textual representations while keeping the core pre-trained T2I models frozen. This significantly expedites the personalization process. For instance, in InstantID \cite{wang2024instantid}, an IdentityNet is designed to integrate facial images with textual prompts, successfully steering image generation in various styles using just a single facial image.

\noindent\textbf{Multi-Concept Customization.}
Existing methods conduct joint training on multi-concept datasets with additional losses or extra optimization efforts to merge multiple models.
Several approaches \cite{avrahami2023break, han2023svdiff, liu2023cones,gong2023talecrafter} employ cross-attention maps to avoid the entanglement of multiple concepts. In Custom Diffusion \cite{kumari2023multi}, the proposition involves joint training or constrained optimization of multiple models. Notably, the work \cite{gu2023mix} introduces gradient fusion to minimize identity loss during concept fusion, along with the proposal of regionally controllable sampling to address attribute binding in multi-concept personalization. Modular Customization \cite{po2023orthogonal} disentangles customization concepts into orthogonal directions, streamlining the integration of multiple fine-tuned concepts, while preserving the integrity of each concept.
\cite{xiao2023fastcomposer} employs subject embeddings from an image encoder to enhance generic text conditioning in diffusion models. This augmentation empowers personalized image generation without the necessity for additional training when facing new concepts.

In contrast to the aforementioned methods, our approach diverges by obviating the need for extensive pre-training of additional network models or the optimization required for merging multiple models.
Through a simple modification of the sampling process, our method seamlessly integrates multiple concepts into a single image using multiple models, thereby eliminating the necessity for model merging or additional tuning.
Furthermore, our method exhibits robust generalization and can be effortlessly combined with various single-concept methods, such as LoRA \cite{hu2021lora} and InstantID \cite{wang2024instantid}, in a plug-and-play manner.

\section{Method}
\label{sec:method}

\begin{figure*}[t]
    \centering
    \includegraphics[width=0.85\textwidth]{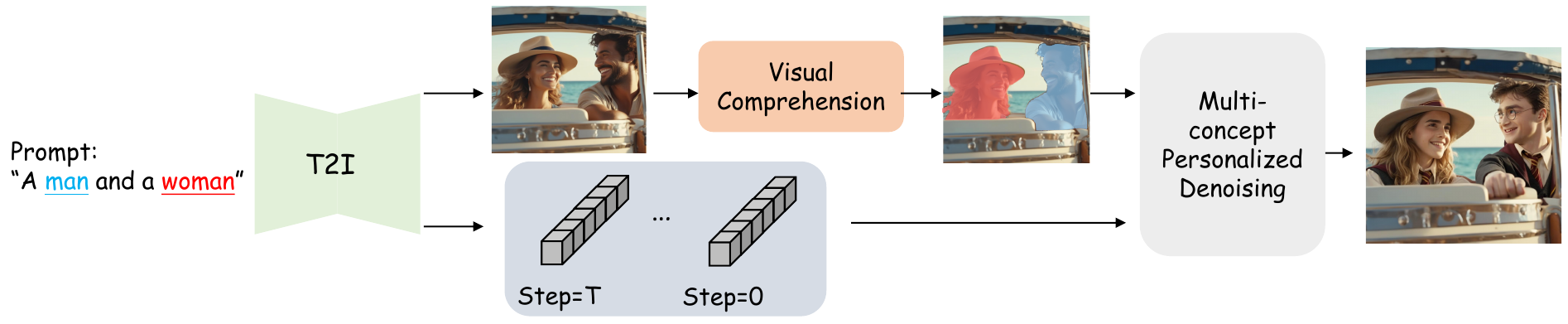}
    \caption{Overviews of the proposed OMG, which contains two stages during sampling. 
    The first stage takes charge of layout generation and visual comprehension information for handling occlusions. Leveraging the acquired information, the identities of concepts can be injected in multi-concept personalized denoising with the proposed latent-level and attention-level noise blending in the second stage. }
    \label{fig:pipeline}
\end{figure*}

We propose a two-stage multi-concept customization framework to integrate multiple concepts into a single image. Unlike previous works, the proposed method can address identity degradation, occlusion, time-consuming fusion, and illumination disharmony problems. 
The overall framework of our proposed paradigm is illustrated in Fig.~\ref{fig:pipeline}, which contains two stages during sampling. 

\subsection{Preliminary}
\label{sec:method-preliminary}

Latent diffusion model \cite{ho2020denoising,song2020denoising,rombach2022high} belongs to a class of generative models containing a diffusion process and a reverse process in the latent space. In the diffusion process, an image $x$ is firstly projected to latent space by an encoder $\mathcal{E}$: $z_0 = \mathcal{E}(x)$. Then random Gaussian noises are gradually added to the data sample $z_0$ to generate the noisy sample $z_t$ with a predefined noise adding schedule $\alpha_t$ as timestep t:
$q(z_t|z_0) = \mathcal{N}(\sqrt{\bar{\alpha}_t}z_0, (1-\bar{\alpha}_t) I)$,
where $\bar{\alpha}_t = \textstyle \prod_{i=1}^{t}\alpha_i $.
In the reverse process, a U-Net $\varepsilon_\theta $ is trained to directly perform denoising in the latent space. The overall training objective is defined as
\begin{equation}
\mathcal{L} = E_{z_0, \epsilon, t}||\epsilon-\varepsilon_\theta(z_t, t, c)||^2_2,
\end{equation}
where $c$ is the embedding of the conditional text prompt and $z_t$ is a noisy sample of $z_0$ at timestep $t$.

\subsection{Stage 1: Visual Comprehension Information Preparation}
\label{sec:method-stage1}

Existing methods, such as Mix-of-show \cite{gu2023mix}, encounter layout conflict challenges. 
As depicted in Fig.~\ref{fig:motivation} (b), when the regions of two concepts occlude, \cite{gu2023mix} is incapable of generating an image with a coherent layout, resulting in identity degradation and compromise of the concept's integrity.
Given that cross-attention layers are effective in controlling the spatial layout and appearance \cite{hertz2022prompt}, the modification of pixel-to-text interaction within these layers allows for preserving the content and spatial layout of the original image while adhering to the target prompt. By selectively modifying predefined regions in an image using a unique identifier, while maintaining the content and structure of other regions, we can effectively mitigate the challenge of concept occlusion. Hence, the first stage aims to acquire visual comprehension information for multi-concept customization.

\begin{figure*}[t]
    \centering
    \includegraphics[width=0.85\textwidth]{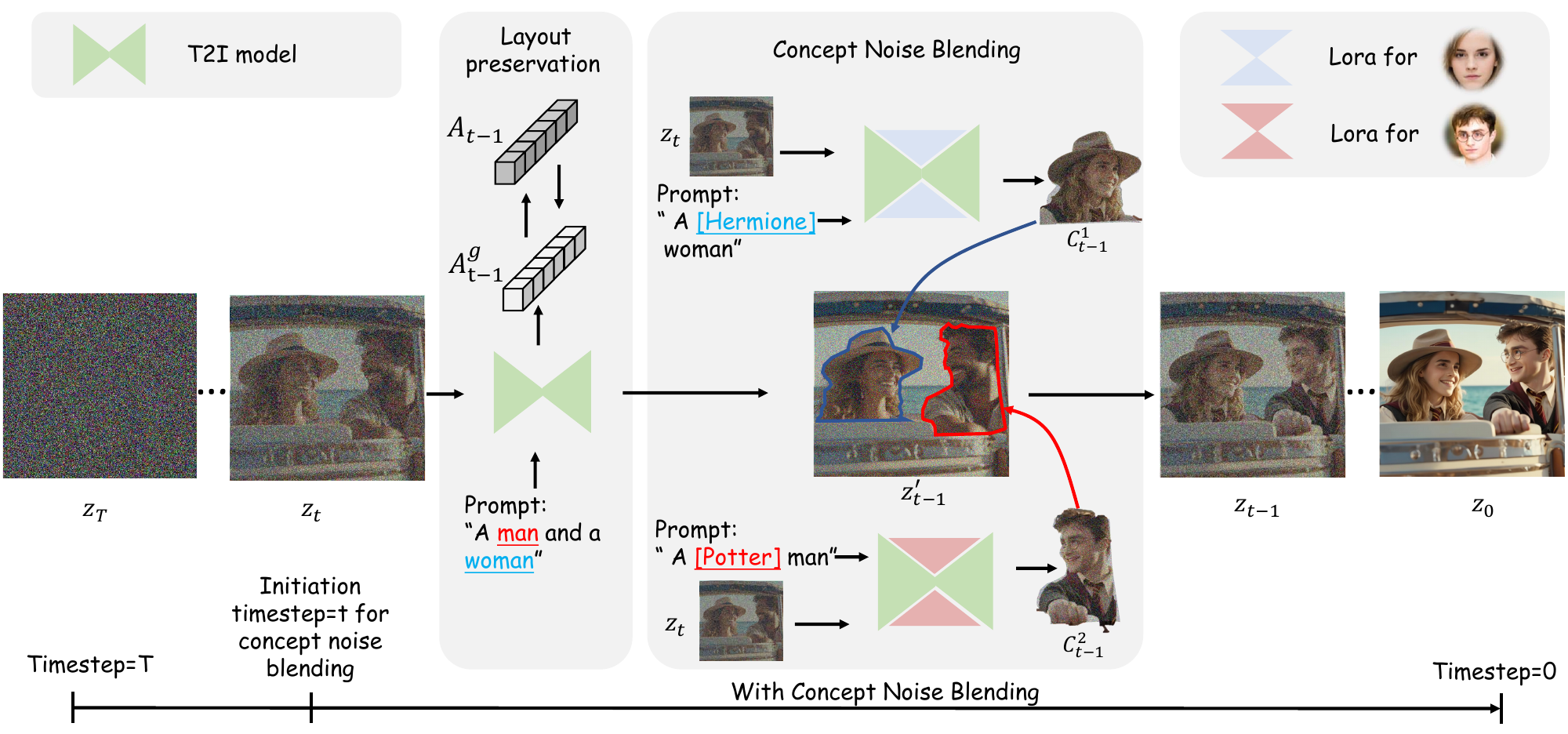}
    \caption{Overviews of the Multi-concept Personalized Denoising. This stage utilizes the acquired visual comprehension information and the designed concept noise blending method to integrate multiple concepts while considering occlusions.}
    \label{fig:blending}
\end{figure*}

As illustrated in Fig.~\ref{fig:pipeline} (a), a textual prompt $p$ describing multiple objects of an image is the input of a T2I model. It is imperative to emphasize that the text prompt $p$ exclusively contains the class name (\textit{e.g.}, ``man'' or ``woman''), deliberately excluding the introduction of the unique identifier (\textit{e.g.}, ``$[v]$ man'' or ``$[v]$ woman'') at this point. Consequently, a non-customized image $x_{ncus}$ with a coherent layout is generated through
\begin{equation}
x_{ncus} = T2I(p).
\end{equation}
We employ the publicly available SDXL model as our T2I model. The denoising UNet network is composed of self-attention layers followed by cross-attention layers. In the denoising process, the fusion of embeddings from visual and text features occurs through cross-attention layers, generating cross-attention maps for each textual token in the U-Net. 
The cross-attention map $A$ is calculated as
\begin{equation}
A = Softmax(\frac{QK^T}{\sqrt{d}} ).
\end{equation}
Here, $Q$ represents a query matrix projection of intermediate features $\varphi(z_t)$, and $K$ is a key matrix projection of text tokens $\phi(p)$, obtained through two learnable linear projections $W_Q$ and $W_K$, respectively. $Q$ and $K$ are defined as
\begin{equation}
Q = W_Q \cdot \varphi(z_t), K = W_K \cdot \phi (p).
\end{equation}
At each denoising step $t$, following the input of $p$ to the T2I model, the cross-attention maps $A_t$, comprising $N$ attention layers with corresponding spatial attention maps $\{A_t^1, A_t^2, \cdots, A_t^N\}$, are acquired. It is imperative to retain all these obtained attention maps for identity injection in the second stage.

To prepare for concept noise blending, it is necessary to locate the modified region in $x_{ncus}$. 
Relying on the robust image understanding capabilities \cite{kirillov2023segment} of visual comprehension, we can derive concept masks $M$. By inputting both the generated image $x_{ncus}$ and the class name (\textit{e.g.}, ``man'' or ``woman'') from $p$, concept masks $M$ corresponding to $k$ class $\{M_1, M_2, \cdots, M_k\}$ can be derived.


\subsection{Stage 2: Multi-concept Personalized Denoising}
\label{sec:method-stage2}

Upon obtaining a non-customized image $x_{ncus}$ with acquired visual comprehension information, we inject the identity of concepts in the second stage. In previous works, such as \cite{hertz2022prompt}, image editing is achieved by injecting the input text with an edit text prompt. Personalized multi-concept generation could adopt a similar approach by triggering concept generation through the identifiers in text prompts. However, it may face two drawbacks.
Firstly, making a text prompt capable of generating multiple concepts necessitates the merging of multiple single-concept models into one like \cite{gu2023mix, po2023orthogonal}, which requires additional network optimization and is inherently time-consuming. 
Additionally, as illustrated in Fig.~\ref{fig:motivation} (a), employing a single prompt for a multi-concept generation often results in identity degradation. 
In contrast, we propose a Concept Noise Blending strategy to address the aforementioned issues. The overall architecture of the multi-concept personalized denoising is depicted in Fig. \ref{fig:blending}.

\begin{figure*}[t]
    \centering
    \includegraphics[width=0.8\textwidth]{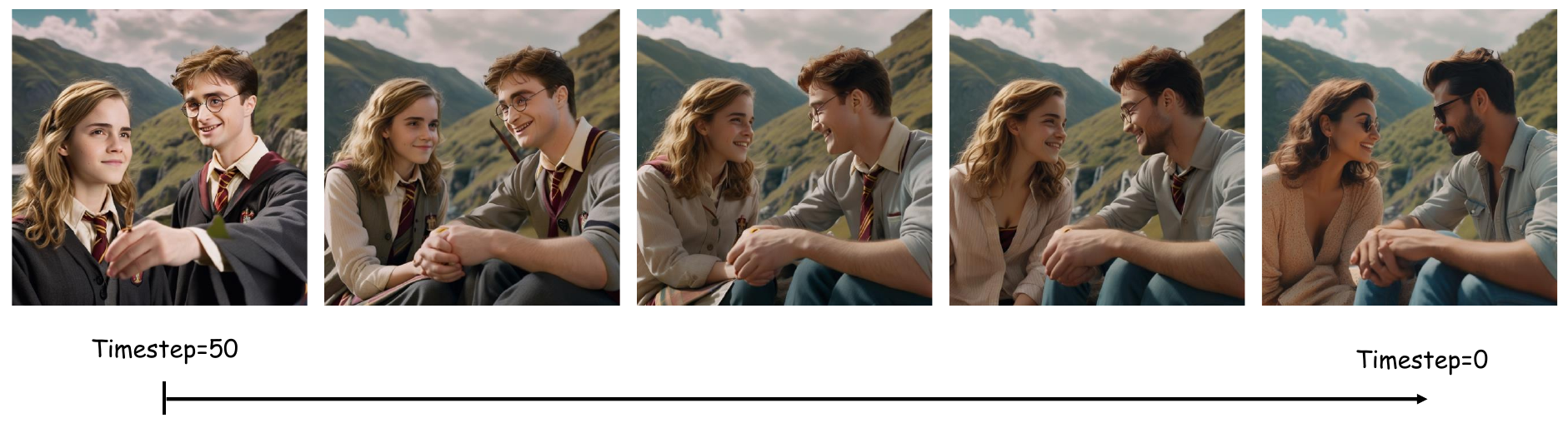}
    \caption{Effect of the initiation timestep for concept noise blending. The initiation timestep for concept noise blending influences both the image layout and illumination. When the initiation timestep is $0$, there is no concept noise blending operation during sampling, resulting in the same generation result for both stages.}
    \label{fig:timestpe}
\end{figure*}

\noindent\textbf{Concept Noise Blending.}
To mitigate the additional optimization costs associated with network merging, the proposed method directly leverages multiple single-concept models during inference, circumventing the need for network merging.
Moreover, each single-concept model is solely responsible for generating a specific concept, effectively addressing the challenge of identity degradation.

During the multi-concept personalized denoising, the input global text prompt $p$ and initiation noise remain consistent at the first stage. 
In the second stage, the objective is to generate a customized image containing multiple concepts leveraging the acquired visual comprehension information. Suppose we aim to generate an image $x_{cus}$ containing $k$ concepts $\{C^1, C^2, \cdots, C^k\}$. Let $T2I^i_c$ represent the $i$-th single-concept model designed to generate the concept $C^i$ through concept text prompt $p^i$. The $p^i$ encapsulates a special identifier that can be input to $T2I^i_c$ for generating concept $C^i$.
At timestep $t$, given text prompt $p^i$ of concept $i$, the corresponding predicted noise $C_{t-1}^i$ is obtained through
\begin{equation}
C_{t-1}^{i} = T2I^i_c(z_{t}, p^i, t).
\end{equation}
Additionally, the $T2I$ model is the same as the first stage. By inputting a global text prompt $p$ at timestep $t$, the corresponding global output $z_{t-1}^{'}$ is obtained through the $T2I$ model with occlusion layout preservation. 
The generated $z_{t-1}^{'}$ represents a non-customized noise. To inject the identity of the concept $C^i$ into $z_{t-1}^{'}$, specific regions in $z_{t-1}^{'}$ are overwritten with the corresponding concept noise $C_{t-1}^{i}$ based on concept masks $M$ through:
\begin{equation}
z_{t-1} = (1- {\textstyle \bigcup_{i=0}^{k}M_i} ) * z_{t-1}^{'} + \sum_{i=0}^{k}{M_i * C_{t-1}^i},
\label{eq:blending}
\end{equation}
where $M_i$ denotes the mask for concept $C^i$. Through noise-level concept blending, the identity of concepts can be injected into one noise at each timestep.

MultiDiffusion \cite{bar2023multidiffusion} similarly incorporates noise fusion during sampling, by binding together multiple diffusion generation processes with a shared set of parameters or constraints to generate high-quality and diverse images that adhere to user-provided controls. In contrast, the proposed Concept Noise Blending does not necessitate multiple crops. Instead, different regions are calculated by distinct models. Ultimately, the results from various regions are fused based on the concept mask, eliminating the need for additional optimization steps.

\begin{figure*}[t]
    \centering
    \includegraphics[width=.85\textwidth]{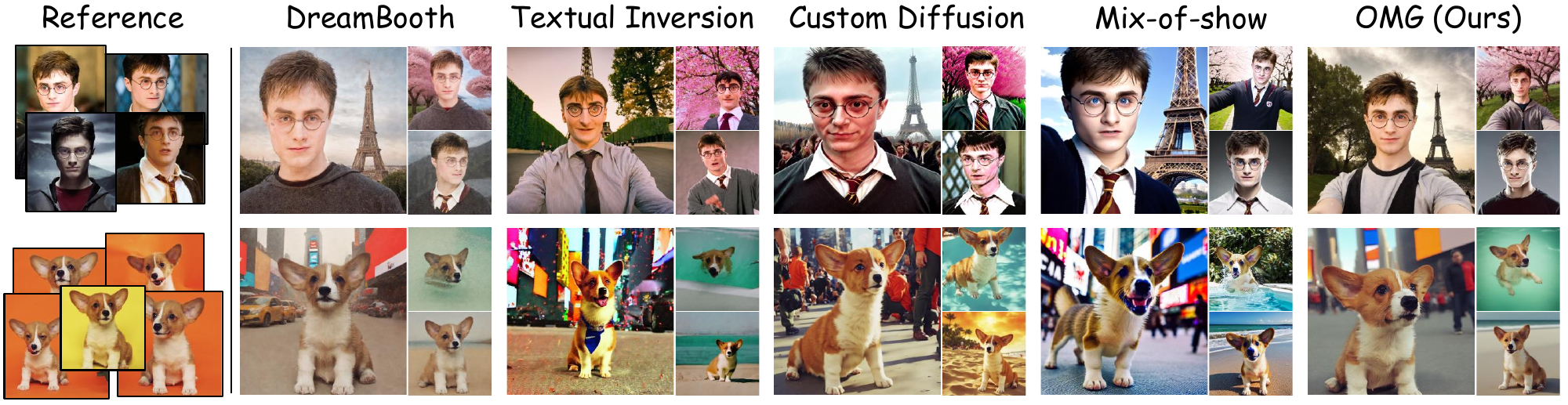}
    \caption{Comparison of OMG with other methods on the single-concept customization. In both character customization and object customization, OMG exhibits superior identity alignment with reference images when compared to other methods.}
    \label{fig:comp_single}
\end{figure*}

\noindent\textbf{Occlusion Layout Preservation.}
The initiation stage yields a non-customized image $x_{ncus}$ with a coherent layout. In the second stage, despite the global prompt and initiation noise being identical to those in the first stage, the generated noises at each timestep are completely different due to Concept Noise Blending. 
We utilize the cross-attention maps $A$ stored in the first stage to uphold the layout consistency of the generated image with $x_{ncus}$. This operation ensures the production of an occlusion-friendly multi-concept customized image.

In each timestep, we ensure that the layout is preserved in the generated image by modifying the cross-attention maps within the UNet during the $T2I$ model sampling. 
For instance, at the $t$ step,
$z_{t}$ is fed into the $T2I$ model alongside the global prompt $p$ and timestep $t$.
Cross-attention maps play a crucial role in controlling the structure and geometry of an image. To maintain an occlusion-friendly layout, we overwrite the generated attention map in each timestep within the UNet with the stored maps. This process can be formulated as:
\begin{equation}
z_{t-1}^{'} = T2I(z_{t}, p, t)\{A^g_{t} \leftarrow A_{t} \},
\end{equation}
where $A^g_{t}$ denotes the generated attention map in the second stage of the $T2I$ model and $A_{t}$ denotes the stored attention map from the first stage.

\subsection{Denoising Timestep of Concept Noise Blending}
\label{sec:method-stepcontrolling}

The initiation timestep for concept noise blending holds significant influence over both the image layout and illumination of the generated image. To elucidate the impact of different concept noise blending starting points, we present generated images at various timesteps in Fig.~\ref{fig:timestpe}. Leveraging DDIM, the series comprises a total of $50$ steps, with the leftmost image representing the outcome when Concept Noise Blending begins at step $50$, indicating that concept noise blending operations are active throughout the entire sampling process. The rightmost image represents the result starting from step $0$, indicating that no concept noise blending operations occur during the entire sampling process. Hence, when the concept noise blending operation starts at timestep $0$, the generated image is the same as stage one.

Commencing concept noise blending at an early step may introduce layout conflicts in the composition and shape of objects within the generated image. However, with the increase in concept noise blending steps, the content information becomes more coherent and stable, effectively preserving the identity of the object. After iterative denoising, as the concept noise blending step approaches $0$, the identity of the character diminishes gradually, resulting in a synthesized image resembling the first stage. This highlights the early stage of sampling governs the image layout, while the identity of concepts unfolds in later timesteps. The optimal step of concept noise blending is approximately $35$.

Moreover, we observe that the illumination disharmony between the foreground and background is notable in the earlier steps. With increasing timesteps, the illumination gradually becomes consistent, suggesting a potential association between illumination and image layout information.

\section{Experiments}
\label{sec:experiments}



\begin{figure*}[t]
    \centering
    \includegraphics[width=.80\textwidth]{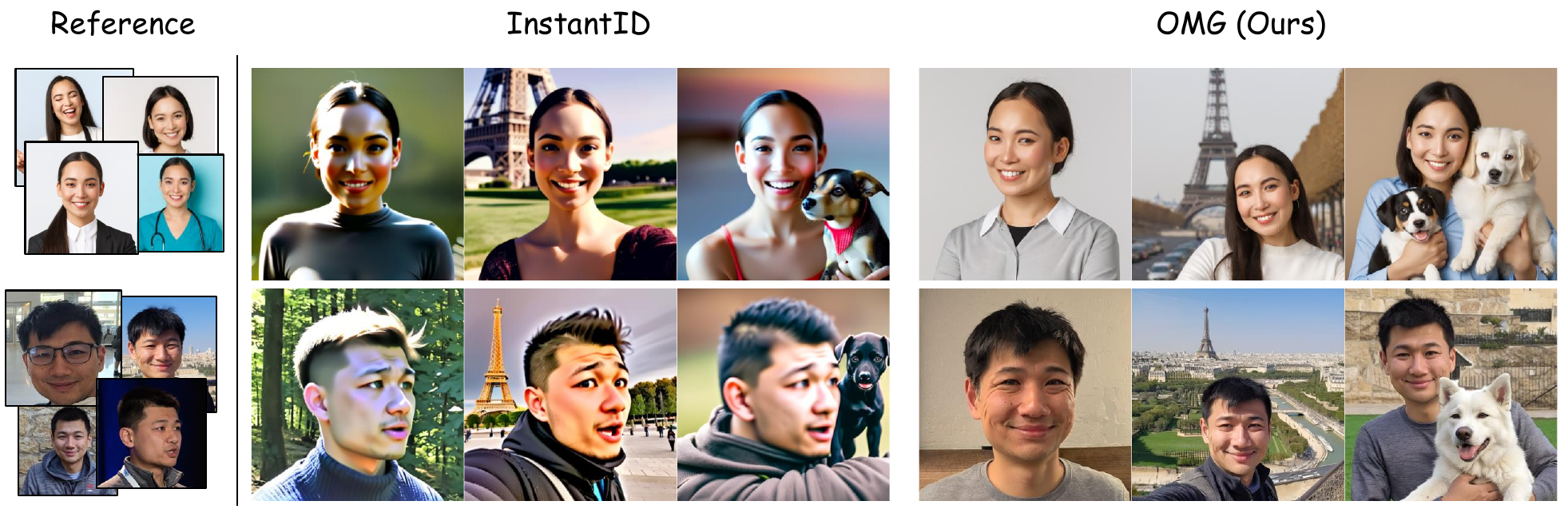}
    \caption{Comparison with InstantID~\cite{wang2024instantid} in single-concept customization. OMG emerges as the superior method by generating images with more natural colors. This showcases the prowess of OMG over InstantID in single-concept customization. For LoRA and InstantID, we all adopt SDXL-base-1.0 as the base model for a fair comparison.}
    \label{fig:instantid-single}
\end{figure*}

\begin{figure*}[t]
    \centering
    \includegraphics[width=.80\textwidth]{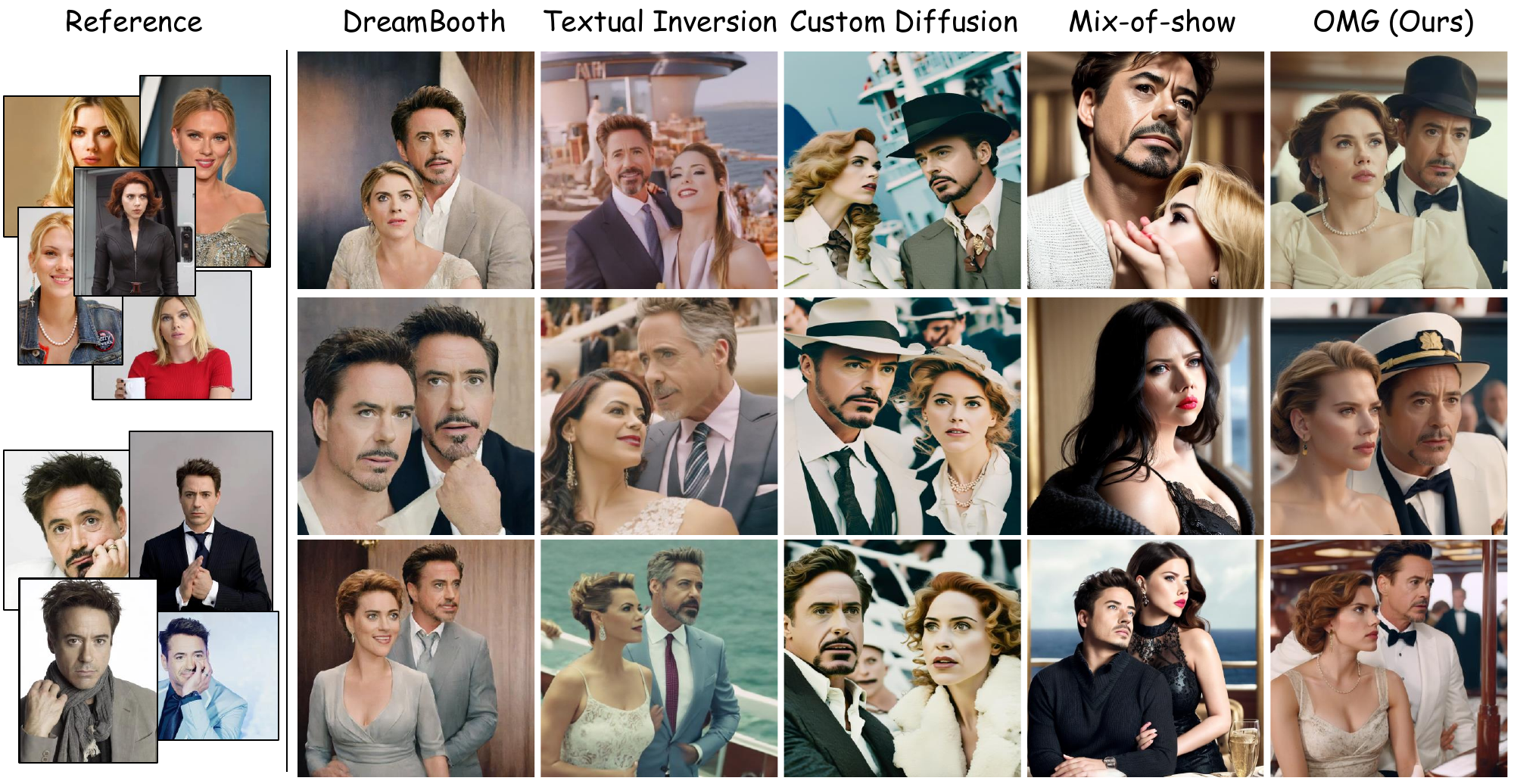}
    \caption{Comparison of OMG with other methods on the same spatial condition on multi-concept customization. To make a fair comparison, all the comparison methods utilize the same spatial condition in each row. The proposed OMG can achieve the best performance in identity preservation in multi-concept customization.}
    \label{fig:comp_multi}
   
\end{figure*}

\begin{figure*}[t]
    \centering
    \includegraphics[width=.70\textwidth]{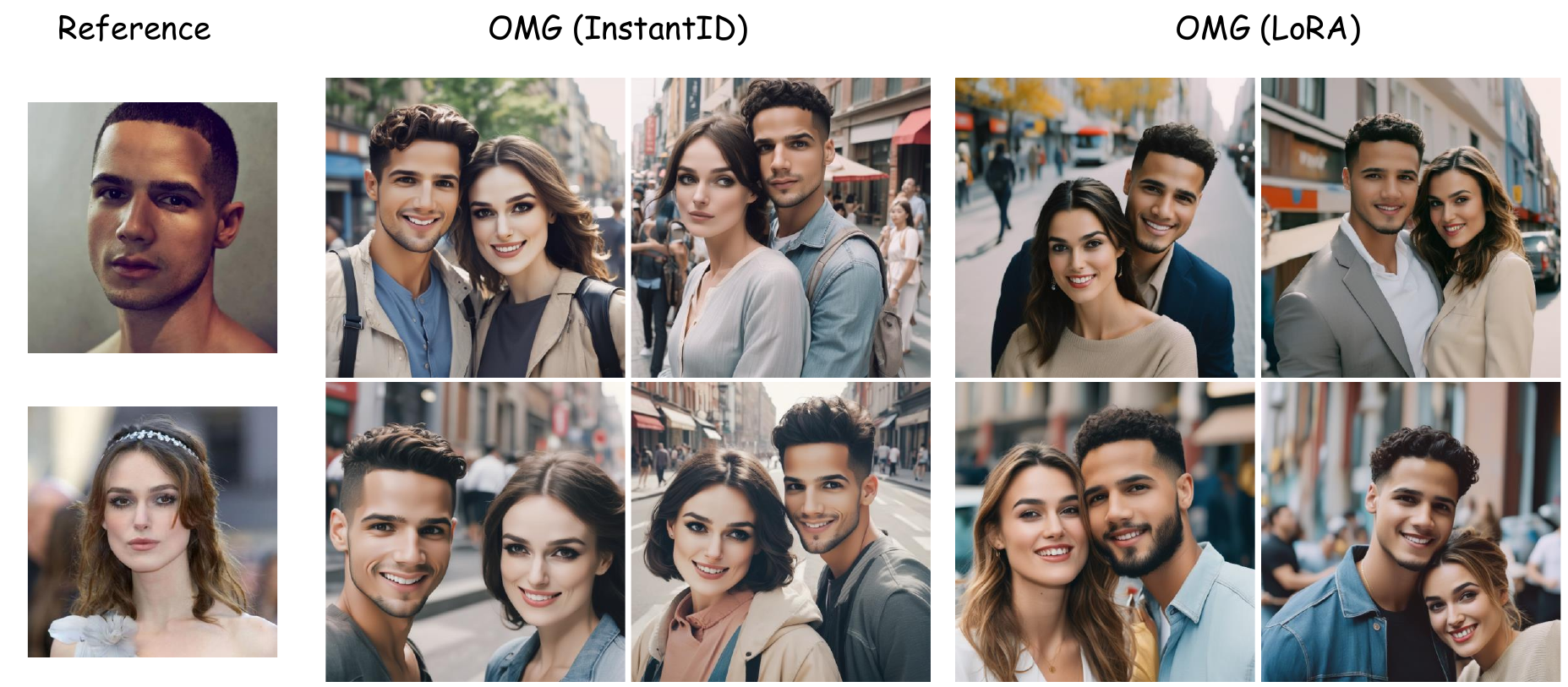}
    \caption{Comparison of OMG with InstantID~\cite{wang2024instantid} in multi-concept customization. OMG stands out by generating images with enhanced realism, characterized by a more extensive and vibrant color spectrum.}
    \label{fig:instantid-multi}
\end{figure*}

\begin{figure*}[t]
    \centering
    \includegraphics[width=.70\textwidth]{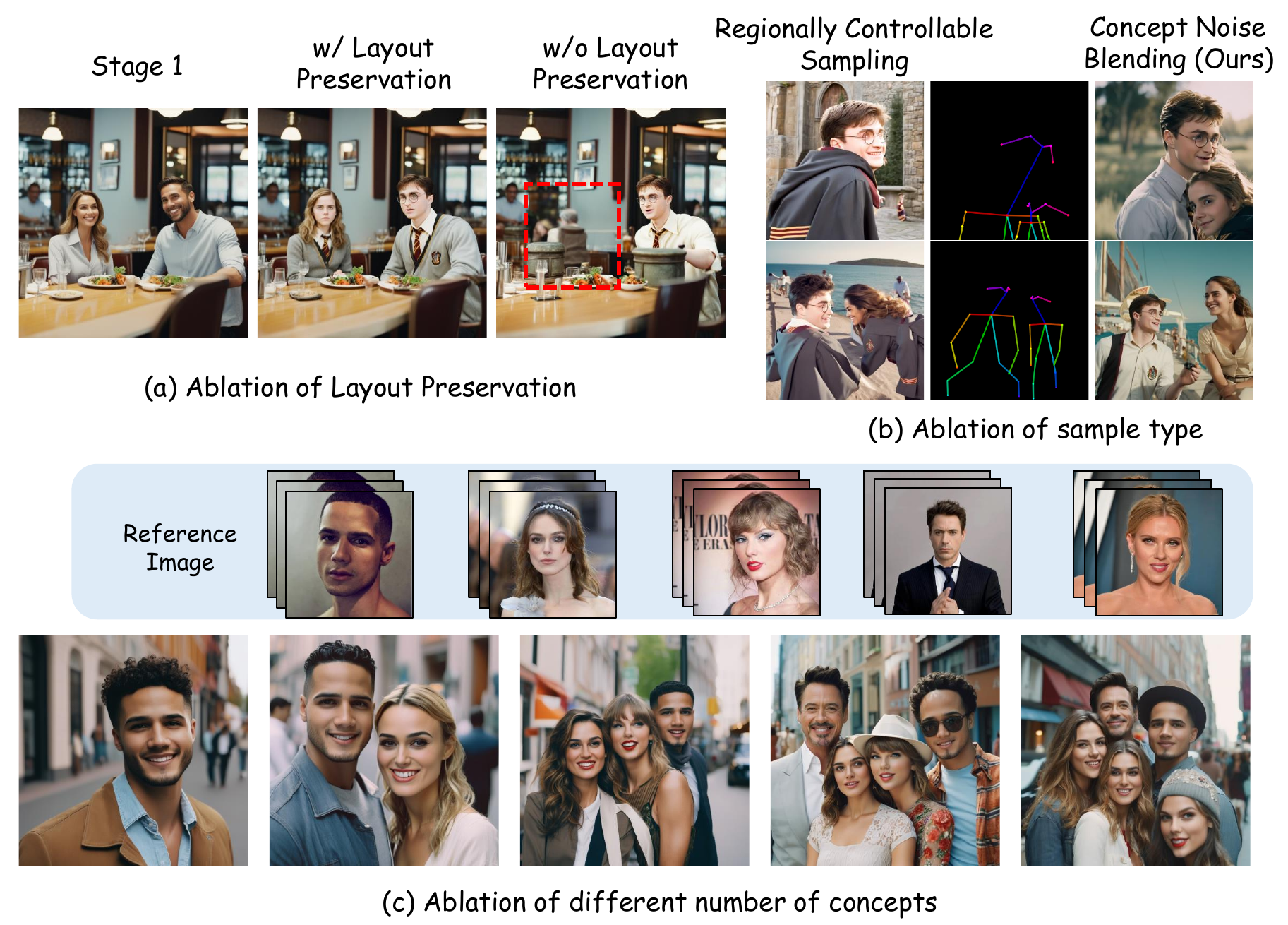}
    \caption{Qualitative ablation study of OMG. (a) Generating images with layout preservation can preserve reasonable structure and enhance realism in the generated images. (b) Concept Noise Blending can generate images with a more coherent image layout and harmonious illumination. (c) The proposed OMG can achieve multi-concept customization with an increasing number of concepts. }
    \label{fig:ablation}
\end{figure*}


\noindent\textbf{Datasets.}
To evaluate OMG method, we collect a dataset that encompasses $15$ distinct concepts. This dataset comprises $7$ real-world characters, $3$ anime characters, and $5$ real-world objects, all annotated automatically by Blip-2 \cite{li2023blip}.

\noindent\textbf{Experimental Setup.}
We implement OMG employing the SDXL model \cite{zhou2023customization}. The multi-concept customization approach we propose can be seamlessly combined with various single-concept customization methods, such as LoRA \cite{hu2021lora} and InstantID \cite{wang2024instantid}.
For LoRA \cite{hu2021lora}, we integrate the LoRA layer into the linear layer in all attention modules of the text encoder and Unet, with a rank of $256$. We use the Adafactor optimizer with a constant learning rate for all experiments, setting the learning rate for the text encoder to $3e^{-5}$ and for Unet to $3e^{-3}$. Single-concept fine-tuning requires approximately $2$ hours on one A100 GPU.
Regarding InstantID \cite{wang2024instantid}, we leverage the officially provided pre-trained Image Adapter model and IdentityNet model. We utilize the Antelopev2 model for face detection and face ID embedding extraction. When combining the proposed method with InstantID for multi-concept customization, only forward inference is needed during concept image generation, without any additional training.

\noindent\textbf{Evaluation Metrics.}
Following \cite{gu2023mix}, we evaluate our method using Image Alignment, which measures the visual similarity of generated images with the target concept using similarity in the CLIP image feature space \cite{kirillov2023segany}. Additionally, we adopt Text Alignment, which measures the similarity of generated images with given prompts using text-image similarity in the CLIP feature space \cite{kirillov2023segany}. However, for face images, Image Alignment may not accurately evaluate the similarity between the generated face and the real face. To address this, we use Identity Alignment to further illustrate the identity-preserving capabilities by measuring the ArcFace score \cite{deng2019arcface} at which the target human identity is detected in a set of generated images. Consequently, we adopt Text Alignment and Image Alignment for objects, and for characters, Text Alignment and Identity Alignment are employed to measure the performance of methods.

\begin{table*}[t]
\centering
\caption{Quantitative comparison on single- and multi-concept personalization. OMG achieves state-of-the-art performance in single-concept customization and achieves better identity preservation than other methods in multi-concept customization.}
\label{tab:compare}
\resizebox{.90\columnwidth}{!}{%
\begin{tabular}{c|cccccc|cccccc}
\hline
\multirow{3}{*}{Method} & \multicolumn{6}{c|}{Character}                                                                               & \multicolumn{6}{c}{Object}                                                                                   \\ \cline{2-13} 
                        & \multicolumn{3}{c|}{Text Alignment}                             & \multicolumn{3}{c|}{Identity Alignment}    & \multicolumn{3}{c|}{Text Alignment}                             & \multicolumn{3}{c}{Image Alignment}        \\ \cline{2-13} 
                        & Single         & Multiple       & \multicolumn{1}{c|}{$\Delta$} & Single         & Multiple       & $\Delta$ & Single         & Multiple       & \multicolumn{1}{c|}{$\Delta$} & Single         & Multiple       & $\Delta$ \\ \hline
DreamBooth \cite{ruiz2023dreambooth}              & 0.677          & 0.658          & \multicolumn{1}{c|}{-0.019}   & 0.456          & 0.480          & 0.025    & 0.713          & 0.717          & \multicolumn{1}{c|}{0.004}    & 0.805          & 0.800          & -0.005   \\
Textual Inversion \cite{gal2022image}       & 0.673          & 0.673          & \multicolumn{1}{c|}{0.000}    & 0.292          & 0.294          & 0.002    & 0.693          & 0.697          & \multicolumn{1}{c|}{0.004}    & 0.784          & 0.781          & -0.003   \\
Custom Diffusion \cite{kumari2023multi}       & 0.629          & \textbf{0.704} & \multicolumn{1}{c|}{0.075}    & 0.370          & 0.322          & -0.048   & 0.695          & 0.755          & \multicolumn{1}{c|}{0.060}    & 0.840          & 0.778          & -0.061   \\
Mix-of-show \cite{gu2023mix}            & 0.675          & 0.639          & \multicolumn{1}{c|}{-0.036}   & 0.422          & 0.436          & 0.015    & 0.724          & 0.731          & \multicolumn{1}{c|}{0.007}    & 0.791          & 0.780          & -0.011   \\ \hline
OMG (Ours)              & \textbf{0.693} & 0.696          & \multicolumn{1}{c|}{0.003}    & \textbf{0.514} & \textbf{0.510} & -0.004   & \textbf{0.730} & \textbf{0.762} & \multicolumn{1}{c|}{0.032}    & \textbf{0.842} & \textbf{0.810} & -0.033   \\ \hline
\end{tabular}%
}
\end{table*}

\subsection{Quantitative Comparison}

We compare OMG with several concept customization methods, including DreamBooth~\cite{ruiz2023dreambooth}, Textual Inversion~\cite{gal2022image}, InstantID~\cite{wang2024instantid}, Custom Diffusion~\cite{kumari2023multi}, and Mix-of-show~\cite{gu2023mix}. 
All the methods except InstantID are training-based customization requiring multiple reference images. In contrast, InstantID~\cite{wang2024instantid} achieves personalized generation with just one reference image.

Following Custom Diffusion~\cite{kumari2023multi}, we utilize $20$ text prompts and $50$ samples per prompt for each concept. Hence, a total of $1000$ images are ultimately generated. For a fair comparison, all the comparison methods adopt DDIM sampling with $50$ steps and a classifier-free guidance sample across all methods.
Our evaluation spans various categories of concepts, including characters and objects. We use a single-concept tuned model to assess the identity-preserving effect of our method through a set of prompts. The experimental results including single-concept and multi-concepts are detailed in Tab.~\ref{tab:compare}. 

For single-concept, we achieve the best results in Text Alignment, Image Alignment, and Identity Alignment for characters and objects. We adopt LoRA for single-concept fine-tuning, which proves the effectiveness of LoRA in capturing the complex concepts’ identity. For multi-concept, the proposed method exhibits superior performance with the input images for object customization. For characters, our method performs better on Identity Alignment than other methods, which proves the superiority of our method in identity preservation.


In our comparative analysis, we compare the proposed method with InstantID~\cite{wang2024instantid}. Notably, InstantID~\cite{wang2024instantid} achieves image customization requiring only a single reference image for inference, while ours leverages multiple reference images for fine-tuning. To ensure an equitable comparison, we align the number of reference images used by InstantID with our approach and calculate the average mean of ID embeddings as an image prompt. Consequently, our method achieves a Text Alignment score of $0.692$ and an Identity Alignment score of $0.500$. InstantID, exhibiting superior performance with a Text Alignment score of $0.698$ and an Identity Alignment score of $0.534$, benefits from fine-tuning on ample facial data. It is notable that our method, in contrast, has not undergone fine-tuning on such extensive datasets.


\begin{table*}[]
\centering
\caption{Quantitative results for diverse concepts generation.}
\label{tab:multi-arc}
\resizebox{.78\columnwidth}{!}{%
\begin{tabular}{c|cccc|c}
\hline
\multirow{2}{*}{Method} & man + woman    & man + man      & woman + woman  & object + object & \multirow{2}{*}{Average} \\ \cline{2-5}
                        & IDA            & IDA            & IDA            & IMA             &                          \\ \hline
DreamBooth \cite{ruiz2023dreambooth}             & 0.302          & 0.258          & 0.192          & \textbf{0.784}  & 0.384                    \\
Textual Inversion \cite{gal2022image}      & 0.122          & 0.131          & 0.064          & 0.675           & 0.248                    \\
Custom Diffusion \cite{kumari2023multi}       & 0.265          & 0.210          & 0.212          & 0.757           & 0.361                    \\
Mix-of-show \cite{gu2023mix}            & 0.361          & 0.219          & 0.143          & 0.736           & 0.365                    \\ \hline
OMG (Ours)                    & \textbf{0.487} & \textbf{0.377} & \textbf{0.293} & 0.763           & \textbf{0.480}           \\ \hline
\end{tabular}%
}
\end{table*}

The qualitative results for multi-concept personalization shown in Tab.~\ref{tab:compare} mainly measure the fusion ability of multiple single-concept models during multi-concept customization. It cannot reflect the generation effect of multi-concept generation. Therefore, to measure the generation effects when different methods generate multiple concepts simultaneously, we propose a new calculation method. To make a fair comparison, we use the same spatial condition to generate two concepts simultaneously. We calculate the region of different concepts in the image through visual comprehension, then calculate the average Identity Alignment scores (IDA) or Image Alignment score (IMA) for two different concepts with their corresponding reference images separately. This measure approach is more effective in measuring the effects of generating multiple concepts simultaneously. 
The experiment results are shown in Tab.~\ref{tab:multi-arc}. OMG outperforms other methods in generating images with various concept combinations, demonstrating the effectiveness of our method in multi-concept generation.

\subsection{Qualitative Comparison}

\subsubsection{Single-Concept Results.}

The efficacy of our method in preserving identity is demonstrated through a comparison of single-concept generation representing different identities. As previously mentioned, each concept undergoes individual fine-tuning. The experimental results are presented in Fig.~\ref{fig:comp_single}. Each column corresponds to images sampled from the same model, representing two 
distinct concept identities. In both character customization and object customization, our method exhibits superior identity alignment with reference images when compared to other methods. The text prompts can be found in the supplement.

Fig. \ref{fig:instantid-single} illustrates the results of single-concept customization compared to InstantID~\cite{wang2024instantid}. Our method stands out by generating higher-quality images, underscoring its visual superiority over InstantID~\cite{wang2024instantid} in single-concept customization.

\subsubsection{Multi-Concept Results.}

We take a comprehensive comparison with other methods in multi-concept customization. Owing that the Mix-of-show~\cite{gu2023mix} requires additional spatial conditions, we implement identical spatial condition controls across all compared methods to make a fair comparison. The experimental results are illustrated in Fig. \ref{fig:comp_multi}. Mix-of-show \cite{gu2023mix} generates layout conflict images, leading to object loss and identity degradation. Notably, DreamBooth~\cite{ruiz2023dreambooth}, Textual Inversion~\cite{gal2022image}, and Custom Diffusion~\cite{kumari2023multi} exhibit limitations in generating images with realistic identity preservation.
In contrast, our proposed method demonstrates robust identity preservation for each character in the multi-concept generation, substantiating its efficacy in multi-concept customization.

Furthermore, we conduct a comparative analysis between the proposed method and InstantID~\cite{wang2024instantid}. To facilitate this comparison, we substitute the single-concept model in our approach with InstantID and juxtapose the two methods. The experimental findings are visually depicted in Fig.~\ref{fig:instantid-multi}. Our method produces images with enhanced realism, with more natural facial. This substantiates the superior performance of our method in multi-concept customization.

\subsection{Ablation Study}

To assess the effectiveness of various components within OMG, we conduct an ablation study encompassing the following elements: Layout Preservation, Concept Noise Blending, and Different Numbers of Concepts. 

\noindent\textbf{Layout Preservation.}
We present the ablation results of layout preservation in Fig.~\ref{fig:ablation} (a). The left image showcases the generated image in the first stage. The other two images illustrate the generated image with and without layout preservation, respectively.
By substituting the attention maps generated in the second stage, the layout of the image is well-preserved. 
The inclusion of layout preservation contributes to the generation of a more reasonable structure, highlighting the effectiveness of layout preservation in enhancing the overall quality.

\noindent\textbf{Concept Noise Blending.}
Subsequently, we compare different sample types, specifically regionally controllable sampling \cite{gu2023mix} and the proposed concept noise blending. Given that regionally controllable sampling necessitates additional spatial conditions, we ensure a fair comparison by providing the same poses for both methods. Experimental outcomes are shown in Fig.~\ref{fig:ablation} (b).
In instances of regionally controllable sampling, occluded regions of two concepts may lead to missing concepts or a disorderly image layout in the generated image. In contrast, the concept noise blending is effective when multiple concepts are occluded. Furthermore, our method yields images with more harmonious illumination between the foreground and the background, resulting in a more realistic portrayal.

\noindent\textbf{Different Numbers of Concepts.}
We also assess the robustness by increasing the number of concepts. As depicted in Fig.~\ref{fig:ablation} (c), we showcase the generation effects when the number of concepts varies from $1$ to $5$. Notably, even with an escalation in the number of concepts, our method consistently preserves the identity of each concept. This substantiates the efficacy of our method in generating a diverse array of concepts while maintaining identity integrity.

\section{Conclusion}
\label{sec:conclusion}

We introduce OMG, a personalized generation framework for handling occlusion challenges in the context of generating realistic images for multiple concepts. Leveraging an image editing framework, our method specifically addresses the occlusion problem prevalent in multi-concept generation. The proposed concept noise blending further mitigates identity degradation issues. Experimental results showcase OMG's ability to successfully generate high-quality images even when concepts experience occlusion. Additionally, our method seamlessly integrates with various single-concept customization models without additional training, enhancing its versatility and practicality.

\section*{Acknowledgment}
This work is funded in part by the National Natural Science Foundation of China (Grant No. 62372480), in part by CCF-Tencent Rhino-Bird Open Research Fund (No. CCF-Tencent RAGR20230118), in part by Theme-based Research Scheme (T45-205/21-N) from Hong Kong RGC, in part by Generative AI Research and Development Centre from
InnoHK.


%
%
\bibliographystyle{splncs04}
\bibliography{egbib}

\begin{thebibliography}{10}
\providecommand{\url}[1]{\texttt{#1}}
\providecommand{\urlprefix}{URL }
\providecommand{\doi}[1]{https://doi.org/#1}

\bibitem{alaluf2023neural}
Alaluf, Y., Richardson, E., Metzer, G., Cohen-Or, D.: A neural space-time representation for text-to-image personalization. ACM TOG  \textbf{42}(6),  1--10 (2023)

\bibitem{arar2023domain}
Arar, M., Gal, R., Atzmon, Y., Chechik, G., Cohen-Or, D., Shamir, A., H.~Bermano, A.: Domain-agnostic tuning-encoder for fast personalization of text-to-image models. In: SIGGRAPH Asia 2023 Conference Papers. pp. 1--10 (2023)

\bibitem{avrahami2023break}
Avrahami, O., Aberman, K., Fried, O., Cohen-Or, D., Lischinski, D.: Break-a-scene: Extracting multiple concepts from a single image. arXiv preprint arXiv:2305.16311  (2023)

\bibitem{bar2023multidiffusion}
Bar-Tal, O., Yariv, L., Lipman, Y., Dekel, T.: Multidiffusion: Fusing diffusion paths for controlled image generation. arXiv preprint arXiv:2302.08113  (2023)

\bibitem{betker2023improving}
Betker, J., Goh, G., Jing, L., Brooks, T., Wang, J., Li, L., Ouyang, L., Zhuang, J., Lee, J., Guo, Y., et~al.: Improving image generation with better captions. Computer Science. https://cdn. openai. com/papers/dall-e-3. pdf  \textbf{2}(3) (2023)

\bibitem{chae2023instructbooth}
Chae, D., Park, N., Kim, J., Lee, K.: Instructbooth: Instruction-following personalized text-to-image generation. arXiv preprint arXiv:2312.03011  (2023)

\bibitem{changpinyo2021conceptual}
Changpinyo, S., Sharma, P., Ding, N., Soricut, R.: Conceptual 12m: Pushing web-scale image-text pre-training to recognize long-tail visual concepts. In: Proceedings of the IEEE/CVF Conference on Computer Vision and Pattern Recognition. pp. 3558--3568 (2021)

\bibitem{chen2023subject}
Chen, W., Hu, H., Li, Y., Rui, N., Jia, X., Chang, M.W., Cohen, W.W.: Subject-driven text-to-image generation via apprenticeship learning. arXiv preprint arXiv:2304.00186  (2023)

\bibitem{chen2023anydoor}
Chen, X., Huang, L., Liu, Y., Shen, Y., Zhao, D., Zhao, H.: Anydoor: Zero-shot object-level image customization. arXiv preprint arXiv:2307.09481  (2023)

\bibitem{choi2023custom}
Choi, J., Choi, Y., Kim, Y., Kim, J., Yoon, S.: Custom-edit: Text-guided image editing with customized diffusion models. arXiv preprint arXiv:2305.15779  (2023)

\bibitem{deng2019arcface}
Deng, J., Guo, J., Xue, N., Zafeiriou, S.: Arcface: Additive angular margin loss for deep face recognition. In: CVPR. pp. 4690--4699 (2019)

\bibitem{gal2022image}
Gal, R., Alaluf, Y., Atzmon, Y., Patashnik, O., Bermano, A.H., Chechik, G., Cohen-Or, D.: An image is worth one word: Personalizing text-to-image generation using textual inversion. ICLR  (2022)

\bibitem{gal2023designing}
Gal, R., Arar, M., Atzmon, Y., Bermano, A.H., Chechik, G., Cohen-Or, D.: Designing an encoder for fast personalization of text-to-image models. arXiv preprint arXiv:2302.12228  (2023)

\bibitem{gal2023encoder}
Gal, R., Arar, M., Atzmon, Y., Bermano, A.H., Chechik, G., Cohen-Or, D.: Encoder-based domain tuning for fast personalization of text-to-image models. ACM TOG  \textbf{42}(4),  1--13 (2023)

\bibitem{gong2023talecrafter}
Gong, Y., Pang, Y., Cun, X., Xia, M., Chen, H., Wang, L., Zhang, Y., Wang, X., Shan, Y., Yang, Y.: Talecrafter: Interactive story visualization with multiple characters. Siggraph Asia  (2023)

\bibitem{gu2023mix}
Gu, Y., Wang, X., Wu, J.Z., Shi, Y., Chen, Y., Fan, Z., Xiao, W., Zhao, R., Chang, S., Wu, W., et~al.: Mix-of-show: Decentralized low-rank adaptation for multi-concept customization of diffusion models. NIPS  (2023)

\bibitem{han2023svdiff}
Han, L., Li, Y., Zhang, H., Milanfar, P., Metaxas, D., Yang, F.: Svdiff: Compact parameter space for diffusion fine-tuning. arXiv preprint arXiv:2303.11305  (2023)

\bibitem{hao2023vico}
Hao, S., Han, K., Zhao, S., Wong, K.Y.K.: Vico: Detail-preserving visual condition for personalized text-to-image generation. arXiv preprint arXiv:2306.00971  (2023)

\bibitem{he2023data}
He, X., Cao, Z., Kolkin, N., Yu, L., Rhodin, H., Kalarot, R.: A data perspective on enhanced identity preservation for diffusion personalization. arXiv preprint arXiv:2311.04315  (2023)

\bibitem{hertz2022prompt}
Hertz, A., Mokady, R., Tenenbaum, J., Aberman, K., Pritch, Y., Cohen-Or, D.: Prompt-to-prompt image editing with cross attention control. arXiv preprint arXiv:2208.01626  (2022)

\bibitem{ho2020denoising}
Ho, J., Jain, A., Abbeel, P.: Denoising diffusion probabilistic models. NeurIPS  \textbf{33},  6840--6851 (2020)

\bibitem{hu2021lora}
Hu, E.J., Shen, Y., Wallis, P., Allen-Zhu, Z., Li, Y., Wang, S., Wang, L., Chen, W.: Lora: Low-rank adaptation of large language models. ICLR  (2021)

\bibitem{kirillov2023segany}
Kirillov, A., Mintun, E., Ravi, N., Mao, H., Rolland, C., Gustafson, L., Xiao, T., Whitehead, S., Berg, A.C., Lo, W.Y., Doll{\'a}r, P., Girshick, R.: Segment anything. arXiv:2304.02643  (2023)

\bibitem{kirillov2023segment}
Kirillov, A., Mintun, E., Ravi, N., Mao, H., Rolland, C., Gustafson, L., Xiao, T., Whitehead, S., Berg, A.C., Lo, W.Y., et~al.: Segment anything. arXiv preprint arXiv:2304.02643  (2023)

\bibitem{kumari2023multi}
Kumari, N., Zhang, B., Zhang, R., Shechtman, E., Zhu, J.Y.: Multi-concept customization of text-to-image diffusion. In: CVPR. pp. 1931--1941 (2023)

\bibitem{li2023blip}
Li, J., Li, D., Savarese, S., Hoi, S.: Blip-2: Bootstrapping language-image pre-training with frozen image encoders and large language models. arXiv preprint arXiv:2301.12597  (2023)

\bibitem{li2023photomaker}
Li, Z., Cao, M., Wang, X., Qi, Z., Cheng, M.M., Shan, Y.: Photomaker: Customizing realistic human photos via stacked id embedding. arXiv preprint arXiv:2312.04461  (2023)

\bibitem{liu2023cones}
Liu, Z., Zhang, Y., Shen, Y., Zheng, K., Zhu, K., Feng, R., Liu, Y., Zhao, D., Zhou, J., Cao, Y.: Cones 2: Customizable image synthesis with multiple subjects. arXiv preprint arXiv:2305.19327  (2023)

\bibitem{ma2023unified}
Ma, Y., Yang, H., Wang, W., Fu, J., Liu, J.: Unified multi-modal latent diffusion for joint subject and text conditional image generation. arXiv preprint arXiv:2303.09319  (2023)

\bibitem{pang2023cross}
Pang, L., Yin, J., Xie, H., Wang, Q., Li, Q., Mao, X.: Cross initialization for personalized text-to-image generation. arXiv preprint arXiv:2312.15905  (2023)

\bibitem{po2023orthogonal}
Po, R., Yang, G., Aberman, K., Wetzstein, G.: Orthogonal adaptation for modular customization of diffusion models. arXiv preprint arXiv:2312.02432  (2023)

\bibitem{rombach2022high}
Rombach, R., Blattmann, A., Lorenz, D., Esser, P., Ommer, B.: High-resolution image synthesis with latent diffusion models. In: CVPR. pp. 10684--10695 (2022)

\bibitem{ruiz2023dreambooth}
Ruiz, N., Li, Y., Jampani, V., Pritch, Y., Rubinstein, M., Aberman, K.: Dreambooth: Fine tuning text-to-image diffusion models for subject-driven generation. In: CVPR. pp. 22500--22510 (2023)

\bibitem{ruiz2023hyperdreambooth}
Ruiz, N., Li, Y., Jampani, V., Wei, W., Hou, T., Pritch, Y., Wadhwa, N., Rubinstein, M., Aberman, K.: Hyperdreambooth: Hypernetworks for fast personalization of text-to-image models. arXiv preprint arXiv:2307.06949  (2023)

\bibitem{saharia2022photorealistic}
Saharia, C., Chan, W., Saxena, S., Li, L., Whang, J., Denton, E.L., Ghasemipour, K., Gontijo~Lopes, R., Karagol~Ayan, B., Salimans, T., et~al.: Photorealistic text-to-image diffusion models with deep language understanding. NIPS  \textbf{35},  36479--36494 (2022)

\bibitem{schuhmann2021laion}
Schuhmann, C., Vencu, R., Beaumont, R., Kaczmarczyk, R., Mullis, C., Katta, A., Coombes, T., Jitsev, J., Komatsuzaki, A.: Laion-400m: Open dataset of clip-filtered 400 million image-text pairs. arXiv preprint arXiv:2111.02114  (2021)

\bibitem{shi2023instantbooth}
Shi, J., Xiong, W., Lin, Z., Jung, H.J.: Instantbooth: Personalized text-to-image generation without test-time finetuning. arXiv preprint arXiv:2304.03411  (2023)

\bibitem{smith2023continual}
Smith, J.S., Hsu, Y.C., Zhang, L., Hua, T., Kira, Z., Shen, Y., Jin, H.: Continual diffusion: Continual customization of text-to-image diffusion with c-lora. arXiv preprint arXiv:2304.06027  (2023)

\bibitem{song2020denoising}
Song, J., Meng, C., Ermon, S.: Denoising diffusion implicit models. ICLR  (2020)

\bibitem{tewel2023key}
Tewel, Y., Gal, R., Chechik, G., Atzmon, Y.: Key-locked rank one editing for text-to-image personalization. In: ACM SIGGRAPH 2023 Conference Proceedings. pp. 1--11 (2023)

\bibitem{tunanyan2023multi}
Tunanyan, H., Xu, D., Navasardyan, S., Wang, Z., Shi, H.: Multi-concept t2i-zero: Tweaking only the text embeddings and nothing else. arXiv preprint arXiv:2310.07419  (2023)

\bibitem{vinker2023concept}
Vinker, Y., Voynov, A., Cohen-Or, D., Shamir, A.: Concept decomposition for visual exploration and inspiration. ACM TOG  \textbf{42}(6),  1--13 (2023)

\bibitem{voynov2023p+}
Voynov, A., Chu, Q., Cohen-Or, D., Aberman, K.: $ p+ $: Extended textual conditioning in text-to-image generation. arXiv preprint arXiv:2303.09522  (2023)

\bibitem{wang2024instantid}
Wang, Q., Bai, X., Wang, H., Qin, Z., Chen, A.: Instantid: Zero-shot identity-preserving generation in seconds. arXiv preprint arXiv:2401.07519  (2024)

\bibitem{wei2023elite}
Wei, Y., Zhang, Y., Ji, Z., Bai, J., Zhang, L., Zuo, W.: Elite: Encoding visual concepts into textual embeddings for customized text-to-image generation. arXiv preprint arXiv:2302.13848  (2023)

\bibitem{xiao2023fastcomposer}
Xiao, G., Yin, T., Freeman, W.T., Durand, F., Han, S.: Fastcomposer: Tuning-free multi-subject image generation with localized attention. arXiv preprint arXiv:2305.10431  (2023)

\bibitem{yan2023facestudio}
Yan, Y., Zhang, C., Wang, R., Zhou, Y., Zhang, G., Cheng, P., Yu, G., Fu, B.: Facestudio: Put your face everywhere in seconds. arXiv preprint arXiv:2312.02663  (2023)

\bibitem{zhang2023compositional}
Zhang, X.L., Wei, X.Y., Wu, J.L., Zhang, T.Y., Zhang, Z.X., Lei, Z., Li, Q.: Compositional inversion for stable diffusion models. arXiv preprint arXiv:2312.08048  (2023)

\bibitem{zhao2023catversion}
Zhao, R., Zhu, M., Dong, S., Wang, N., Gao, X.: Catversion: Concatenating embeddings for diffusion-based text-to-image personalization. arXiv preprint arXiv:2311.14631  (2023)

\bibitem{zhou2023customization}
Zhou, Y., Zhang, R., Gu, J., Sun, T.: Customization assistant for text-to-image generation. arXiv preprint arXiv:2312.03045  (2023)

\bibitem{zhou2023enhancing}
Zhou, Y., Zhang, R., Sun, T., Xu, J.: Enhancing detail preservation for customized text-to-image generation: A regularization-free approach. arXiv preprint arXiv:2305.13579  (2023)

\end{thebibliography}



\end{document}


\title{Supplementary Materials} 

\titlerunning{OMG: Occlusion-friendly Personalized Multi-concept Generation}



\author{
}

\authorrunning{Zhe. K et al.}


\institute{ }

\maketitle

\appendix

This appendix includes our supplementary materials as follows:
\begin{itemize}
\item Evaluation setting in Sec.~\ref{sec:evaluation}.
\item More qualitative results in Sec.~\ref{sec:qualitative}.
\item Combination with ControlNet in Sec.~\ref{sec:controlnet}.
\item Combination with style LoRAs in Sec.~\ref{sec:style}.
\item More results of layout preservation in Sec.~\ref{sec:ablation}.
\item Comparison of different base models for OMG combined with InstantID in Sec.~\ref{sec:model-instantid}.
\item Comparison of different base models for OMG combined with LoRA in Sec.~\ref{sec:model-lora}.
\item Compare with more single image based methods in Sec.~\ref{sec:comp-single}.
\item Model compression in Sec.~\ref{sec:comp-storage}.
\item Limitation and future work in Sec.~\ref{sec:limitation}.
\end{itemize}

\section{Evaluation Setting}
\label{sec:evaluation}

\begin{figure*}[]
    \centering
    \includegraphics[width=0.9\textwidth]{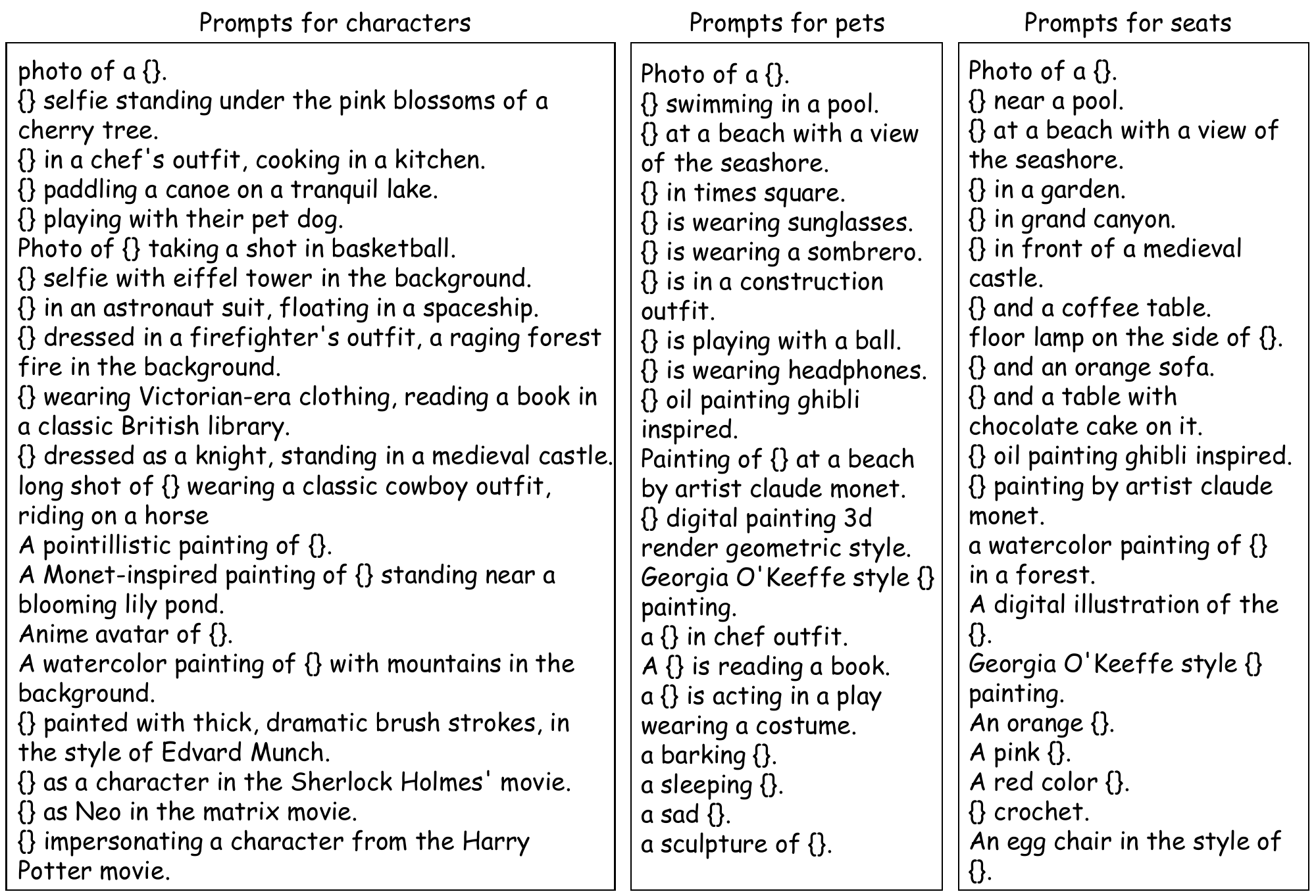}
    \caption{Summary of our evaluation prompts for each concept.}
    \label{fig:prompts}
\end{figure*}

Our dataset for image customization comprises $10$ characters and $5$ objects. For each model, we employ $20$ text prompts, and the evaluation prompts for each concept are presented in Fig. \ref{fig:prompts}.

\section{More Qualitative Results}
\label{sec:qualitative}

\begin{figure*}[t]
    \centering
    \includegraphics[width=0.95\textwidth]{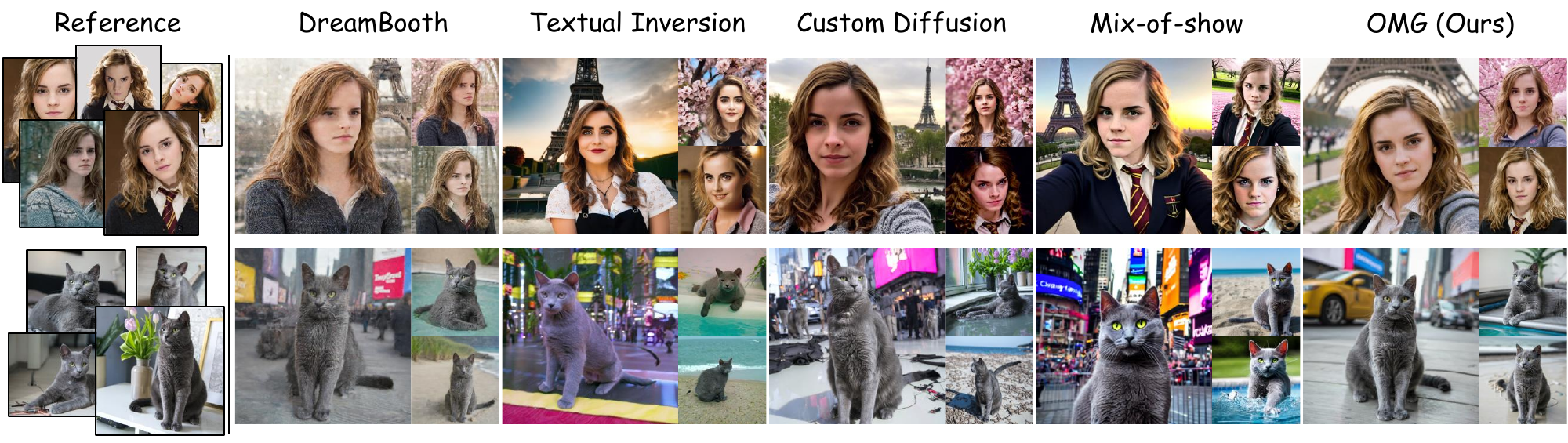}
    \caption{More comparison results of OMG with other methods on the single-concept customization.}
    \label{fig:supp-single}
\end{figure*}

\begin{figure*}[t]
    \centering
    \includegraphics[width=0.9\textwidth]{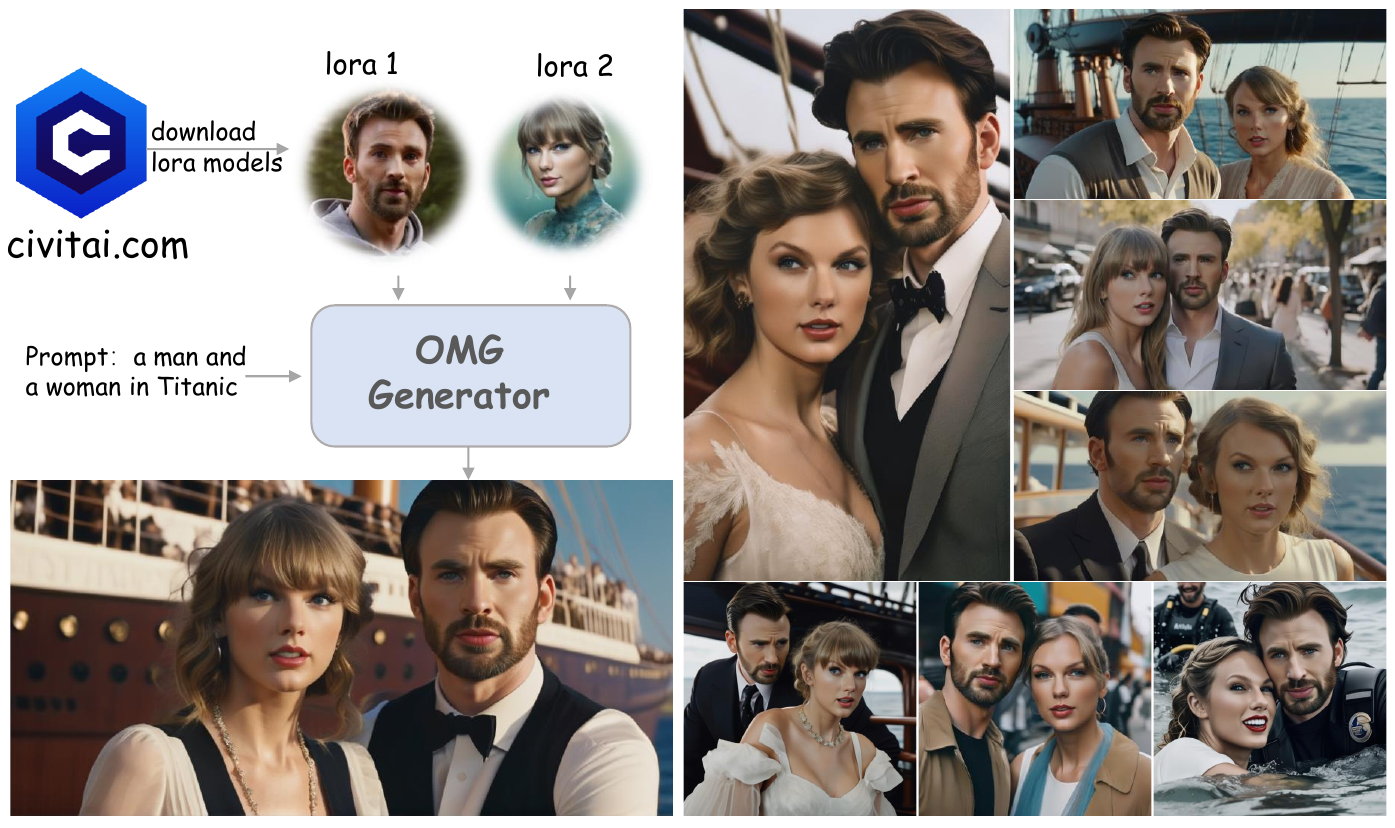}
    \caption{More experimental results of OMG on the multi-concept customization.}
    \label{fig:supp-multi}
\end{figure*}

\begin{figure*}[t]
    \centering
    \includegraphics[width=0.80\textwidth]{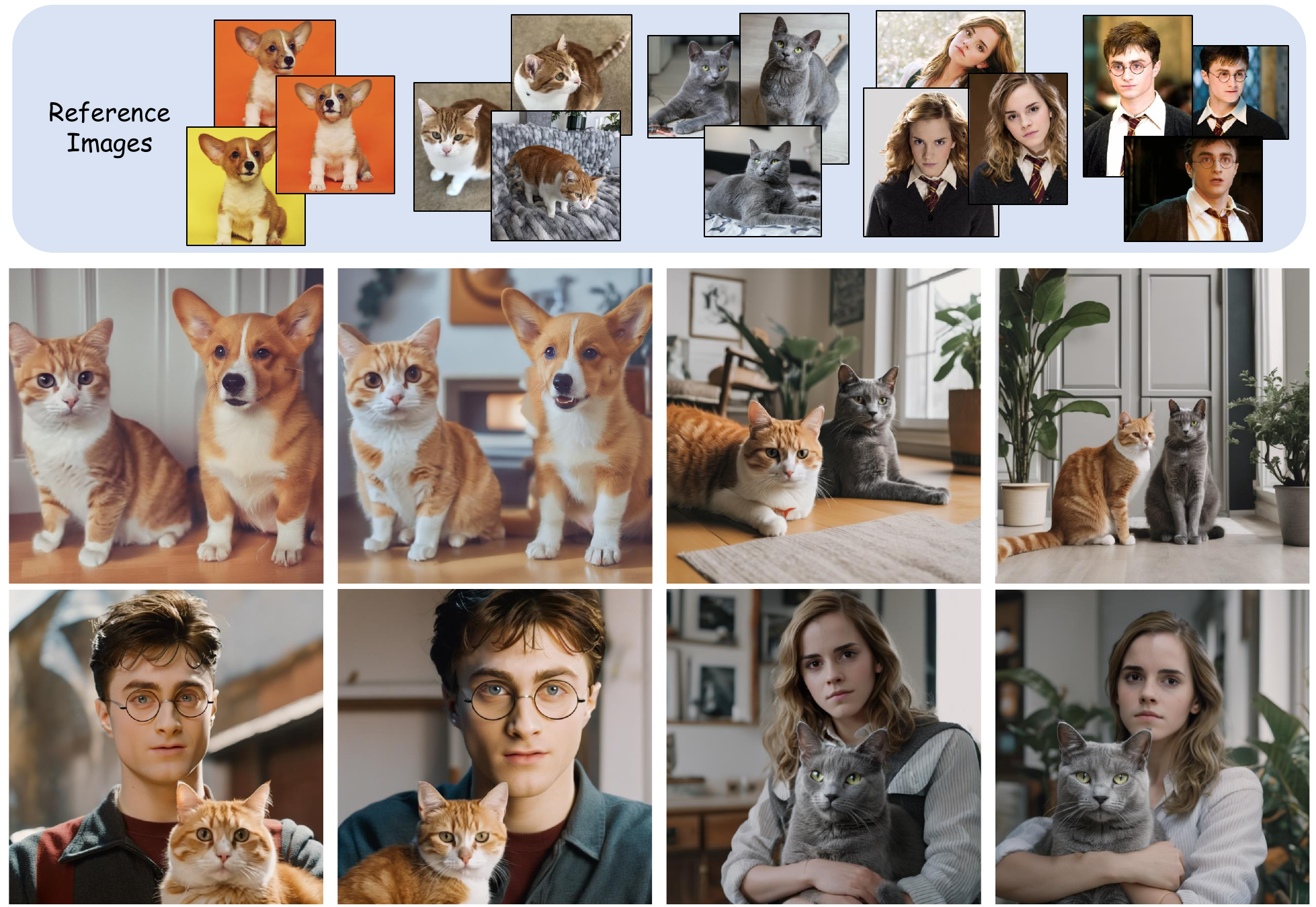}
    \caption{Multi-concept customization results in combining characters and objects.}
    \label{fig:multi-object}
\end{figure*}

We additionally present supplementary experimental results for concept customization. Single-concept customization results compared with other methods are depicted in Fig.~\ref{fig:supp-single}. Multi-concept customization results for characters are shown in Fig.~\ref{fig:supp-multi}. Moreover, further experimental results combining characters and objects are presented in Fig.~\ref{fig:multi-object}.

\section{Combination with ControlNet}
\label{sec:controlnet}

\begin{figure*}[!h]
    \centering
    \includegraphics[width=0.9\textwidth]{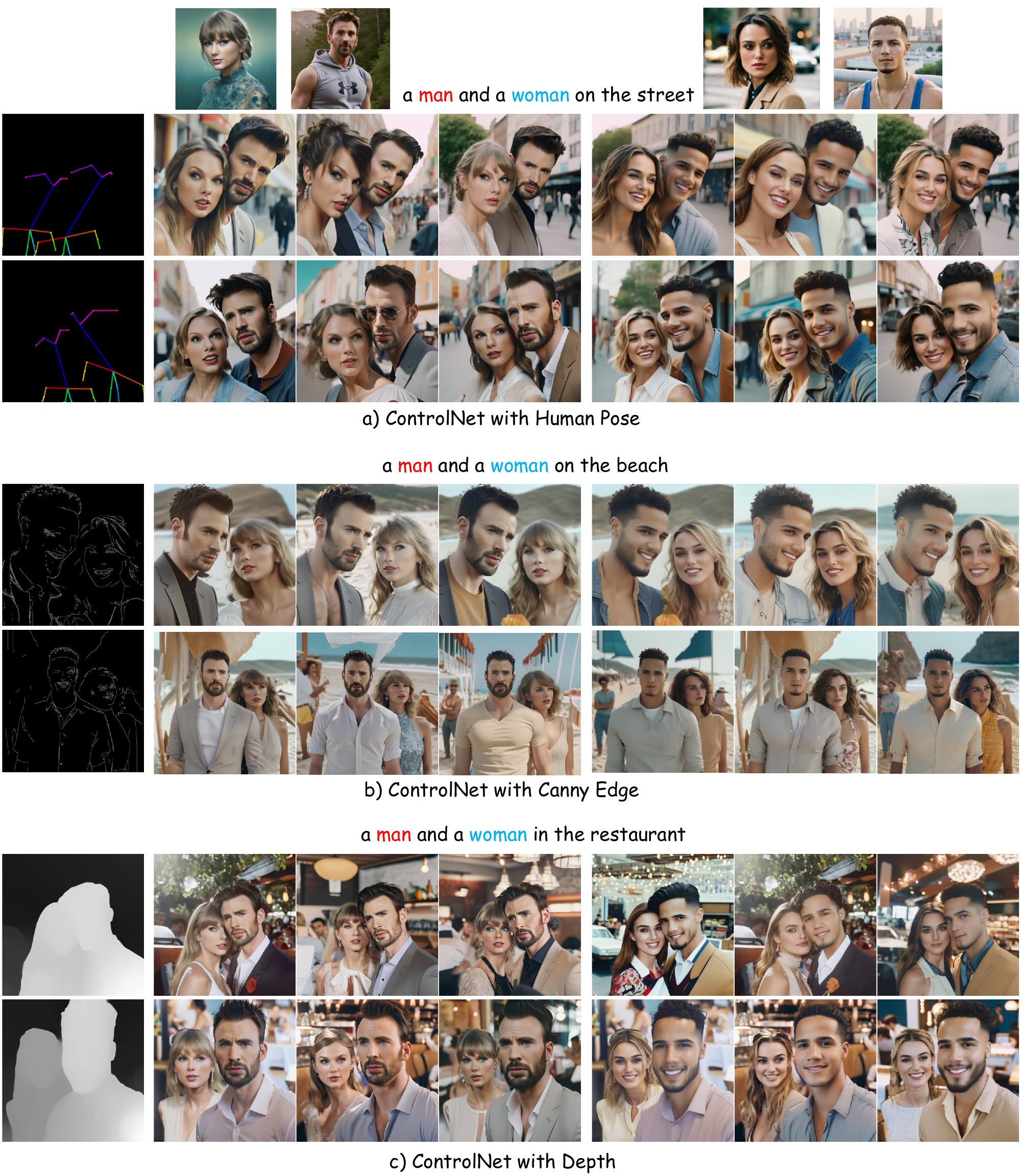}
    \caption{The experimental results of OMG combined with ControlNet under different conditions, including human pose, canny edge, and depth maps.}
    \label{fig:control}
\end{figure*}

The proposed method is versatile and practical, allowing for combination with various conditions using ControlNet. To validate its effectiveness, we conduct experiments by combining ControlNet with different conditions, including human pose, canny edge, and depth maps. The results are presented in Fig.~\ref{fig:control}.

\section{Combination with Style LoRAs}
\label{sec:style}

\begin{figure*}[t]
    \centering
    \includegraphics[width=0.9\textwidth]{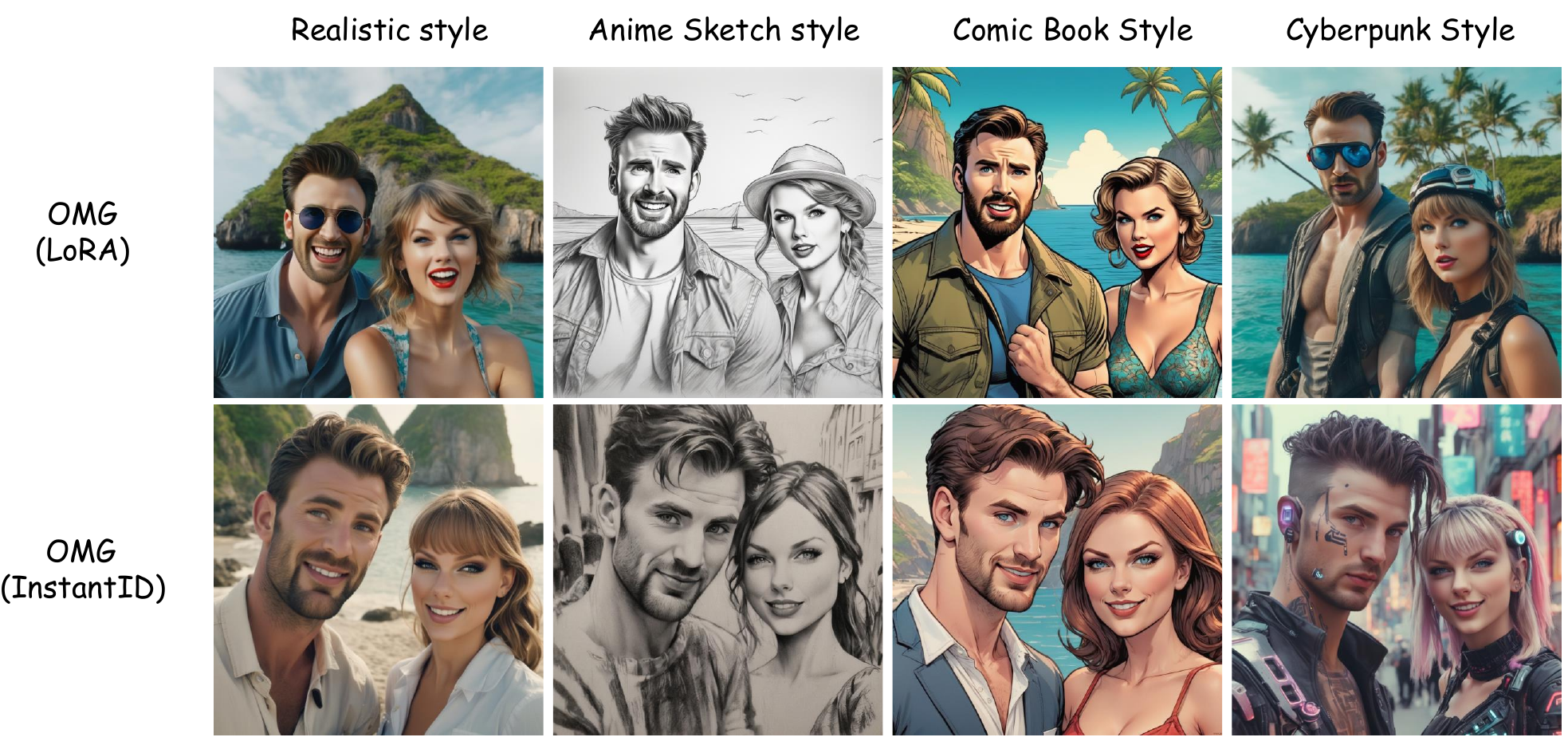}
    \caption{The experimental results of OMG combined with different style LoRAs.}
    \label{fig:style}
\end{figure*}

The proposed method can be combined with different style LoRAs. To validate the effectiveness, we conduct experiments to combine OMG with various LoRAs related to styles, including Anime Sketch style, Comic Book Style, and Cyberpunk Style. We utilize different single-concept customization models, such as LoRA and InstantID. The experimental results are presented in Fig.~\ref{fig:style}, which demonstrate the effectiveness of our method in combining with different styles.

\section{More Results of Layout Preservation}
\label{sec:ablation}

\begin{figure*}[!t]
    \centering
    \includegraphics[width=0.65\textwidth]{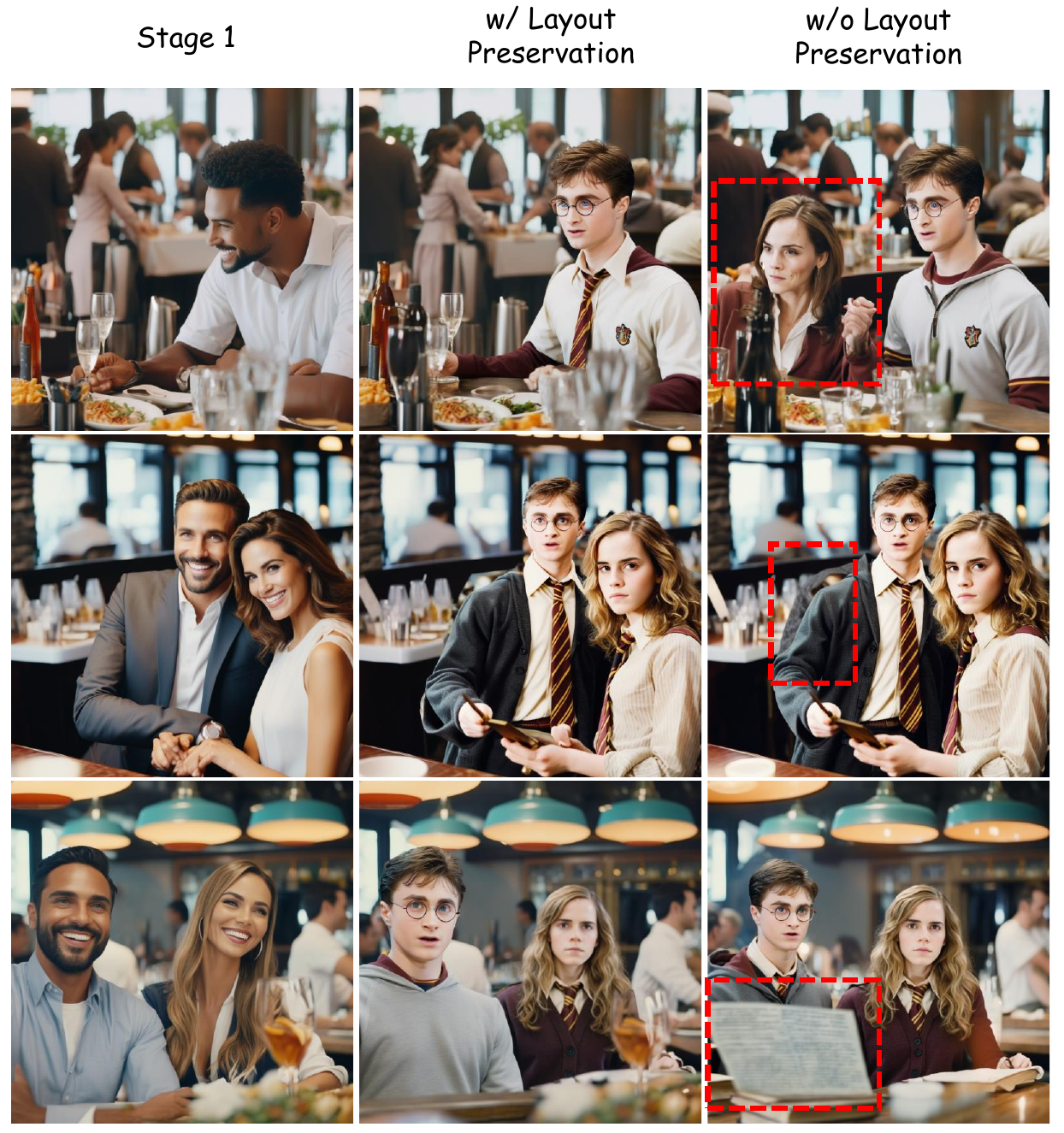}
    \caption{Ablation of layout preservation. Generating images with layout preservation can preserve image structure and enhance image quality.}
    \label{fig:layout}
\end{figure*}

The layout preservation operation in OMG plays a crucial role in maintaining the image layout. To validate the effectiveness of layout preservation, we provide additional examples. As illustrated in Fig.~\ref{fig:layout}, each row showcases the generation results of the first stage, with and without using layout preservation.  The absence of layout preservation can result in the loss of objects or alterations to the overall image structure. This underscores that layout preservation contributes to generating a more reasonable image layout and enhances the overall image quality.

\begin{table*}[t]
\centering
\caption{Quantitative results for ablation study.}
\label{tab:abla_comp}
\resizebox{.5\columnwidth}{!}{%
\begin{tabular}{c|cc|c}
\hline
Method & OMG w RCS & OMG w/o CNB & OMG w CNB           \\ \hline
IDA    & 0.141     & 0.518      & \textbf{0.524} \\ \hline
\end{tabular}%
}
\end{table*}

We also conducted ablation experiments to evaluate the effectiveness of layout preservation and compare concept noise blending (CNB) with regionally controllable sampling (RCS). We calculated the average Identity Alignment scores  (IDA) of multiple concepts. The experimental results are presented in Tab.~\ref{tab:abla_comp}. The results demonstrate that layout preservation is crucial for improving Identity Alignment scores, and concept noise blending is more effective than regionally controllable sampling.

\section{Comparison of different base models for OMG combined with InstantID}
\label{sec:model-instantid}

\begin{figure*}[t]
    \centering
    \includegraphics[width=0.85\textwidth]{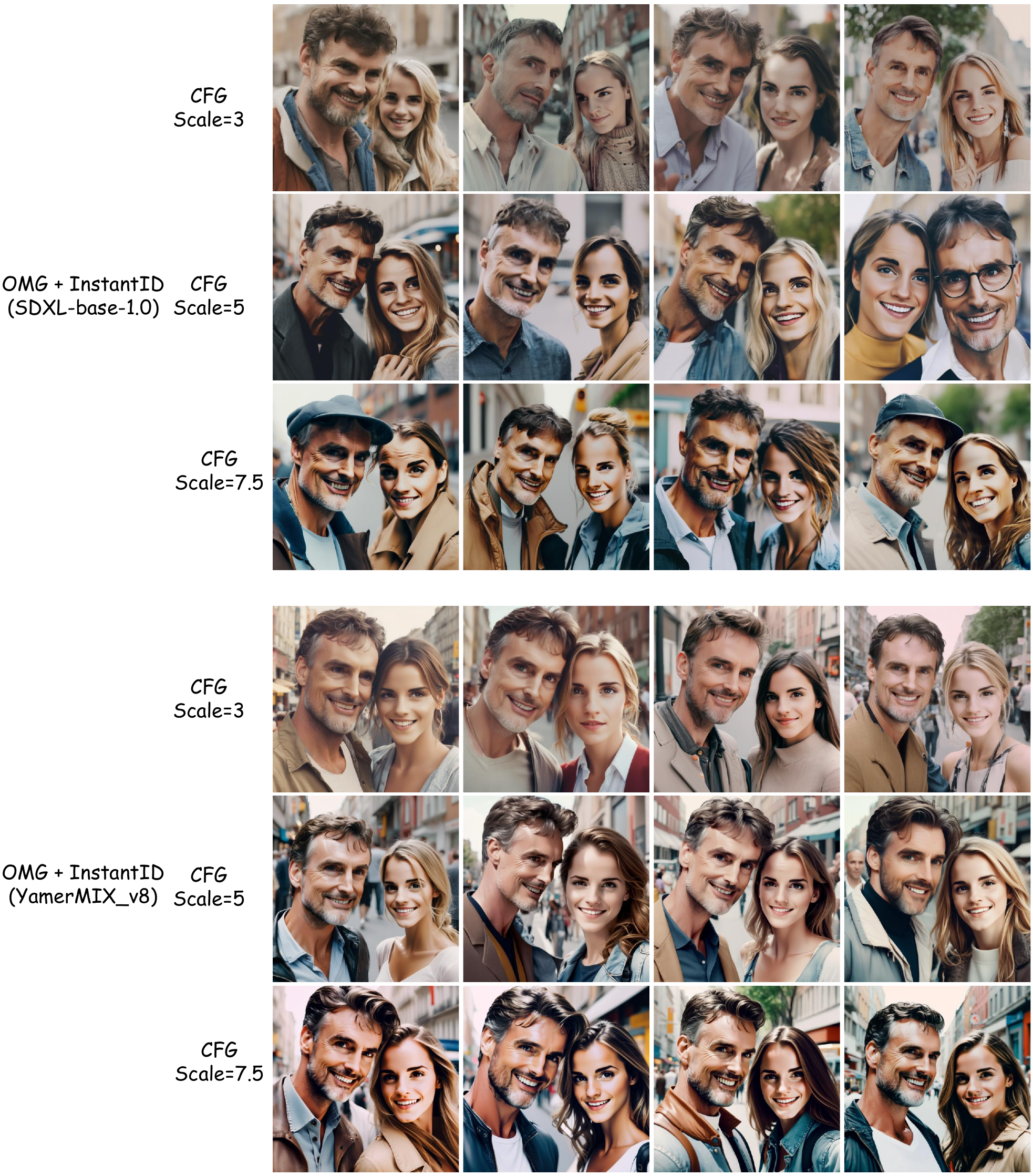}
    \caption{Comparison results of different base models for OMG combined with InstantID.}
    \label{fig:base-model}
\end{figure*}

It's interesting to note the InstantID is sensitivity to the base model and CFG scale. As shown in Fig.~\ref{fig:base-model}, the choice of base model and CFG scale significantly affects the quality of the generated images. Increasing the CFG scale can lead to over-saturation in the results, and the best outcome is achieved when the CFG scale is set to 3. Additionally, using YamerMIX-v8 as the base model seems to help alleviate the over-saturation issue. These findings provide valuable insights for optimizing the InstantID method and achieving better image generation quality.

\section{Comparison of different base models for OMG combined with LoRA.}
\label{sec:model-lora}

\begin{figure*}[t]
    \centering
    \includegraphics[width=0.50\textwidth]{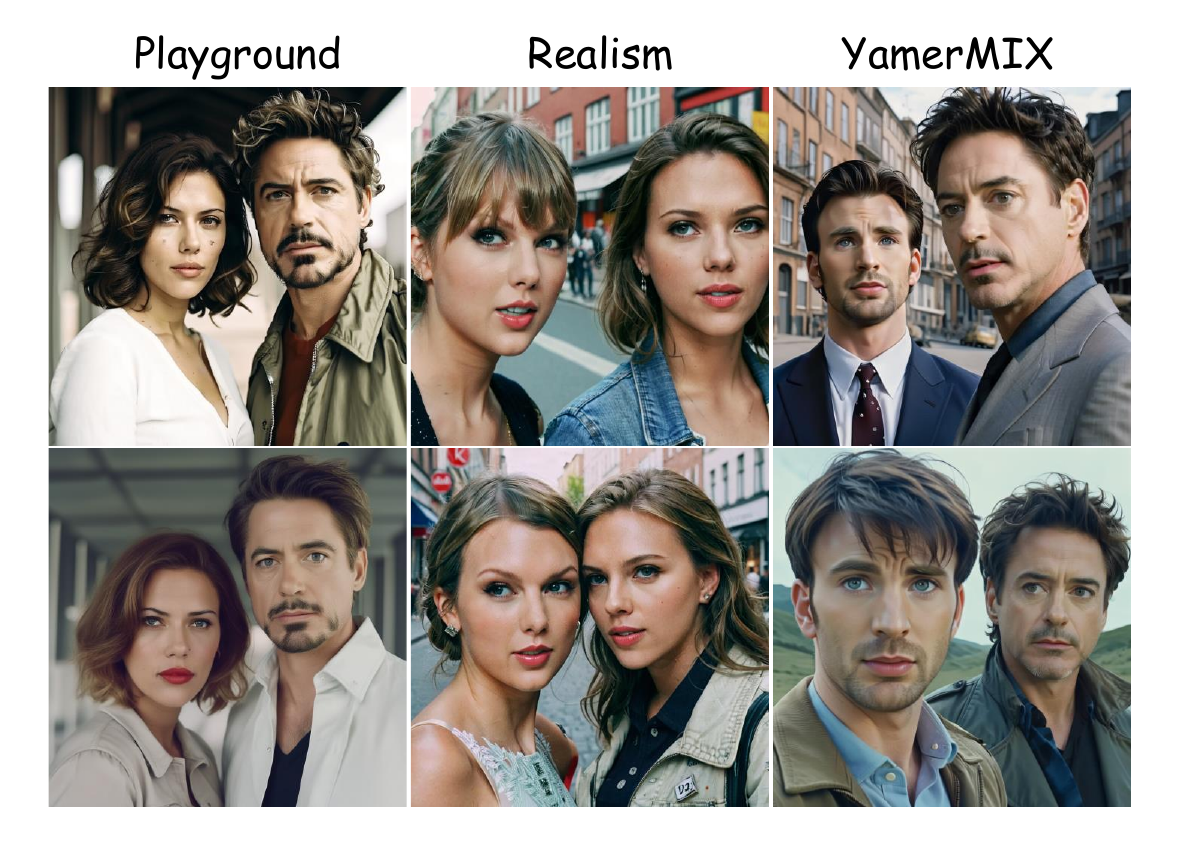}
    \caption{Comparison results of different base models for OMG combined with LoRA}
    \label{fig:base-model-lora}
\end{figure*}

To verify the robustness of OMG combined with LoRA, we conducted experiments on other models, including Realism, YamerMIX-v8, and Playground-v2. The experimental results are presented in Fig.~\ref{fig:base-model-lora}. Our proposed method can generate images with multiple concepts utilizing different base models, where these concepts can belong to the same or different class, demonstrating the robustness of this approach.

\section{Compare with more single image based methods.}
\label{sec:comp-single}

\begin{figure*}[t]
    \centering
    \includegraphics[width=0.78\textwidth]{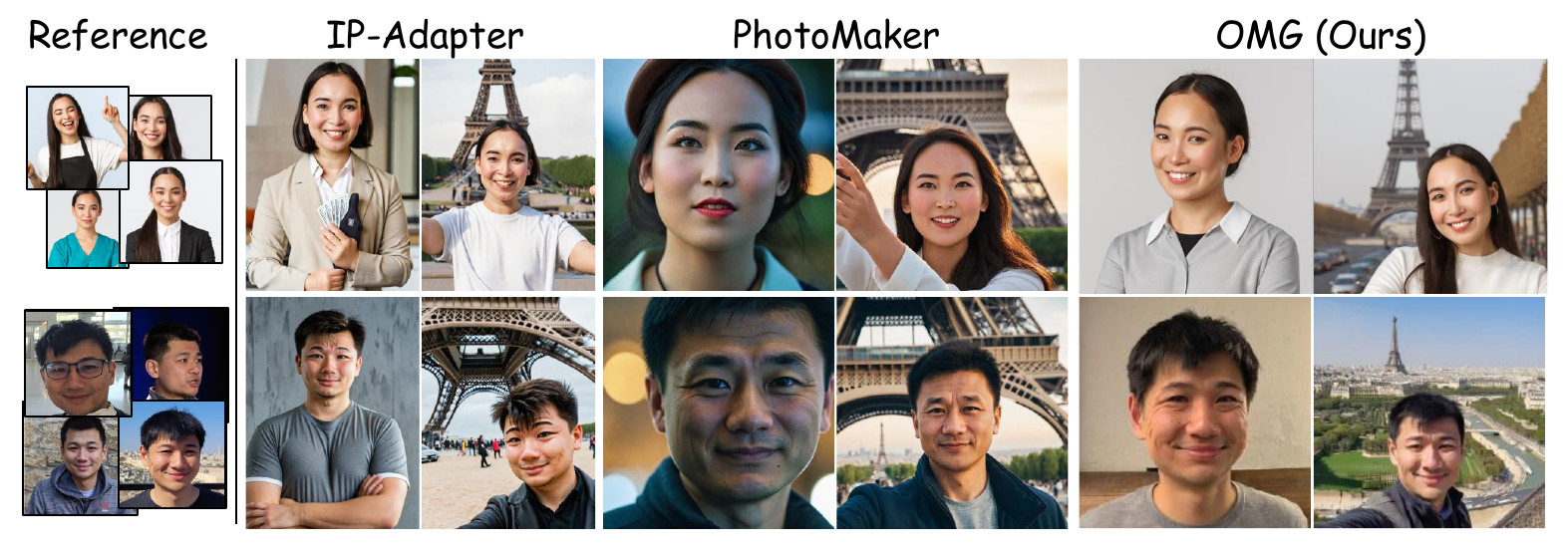}
    \caption{Comparison with more single image based methods.}
    \label{fig:single-image}
\end{figure*}

In addition to InstantID, we also compared our method with other single image based methods, including IP-Adapter and PhotoMaker. The comparison results are presented in Fig.~\ref{fig:single-image}. Our proposed method, OMG, achieves the best identity preservation compared to PhotoMaker and IP-Adapter.

\section{Model compression.}
\label{sec:comp-storage}

\begin{figure*}[t]
    \centering
    \includegraphics[width=0.90\textwidth]{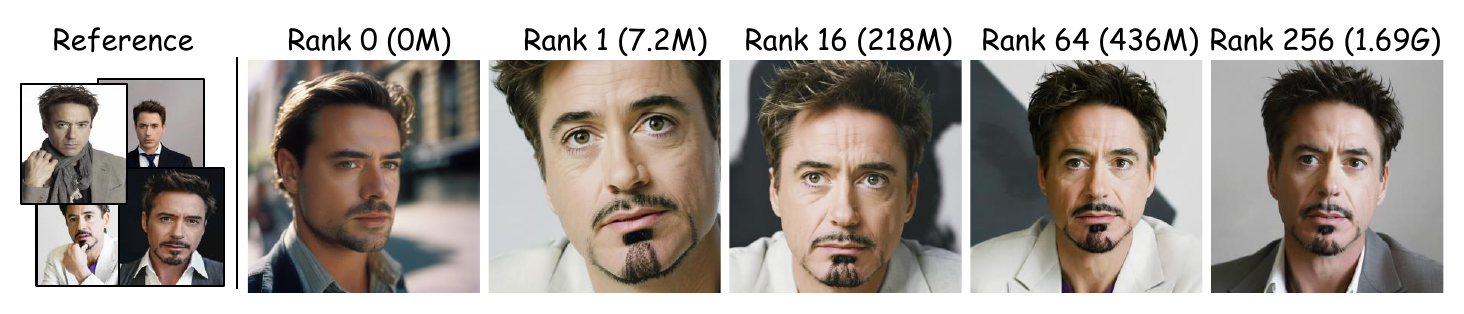}
    \caption{Comparison with different LoRA ranks.}
    \label{fig:ranks}
\end{figure*}

We performed SVD and compared the visualization results of different LoRA ranks, including 0, 1, 16, 64, and 256. As expected, the required storage space increased with the increase of ranks. The visualization results of different ranks are presented in Fig.~\ref{fig:ranks}.

\section{Limitation and Future Work}
\label{sec:limitation}

While our method provides an occlusion-friendly framework for multi-concept personalization with robust identity preservation and harmonious illumination, there are several limitations to consider. Firstly, OMG may face challenges in generating high-quality small-face regions due to information loss in the VAE. Besides, the computational intensity associated with noise fusion from multiple single-concept models are noteworthy consideration, leading to slower generation.

Future research efforts to improve the sample speed of OMG, especially in achieving rapid, high-fidelity sampling, are commendable. The planned exploration of combining OMG with other accelerated methods to generate high-quality images with a few-step inference reflects a proactive approach toward addressing these challenges and enhancing the overall efficiency of the method.
